\documentclass[letterpaper]{article} 
\usepackage{aaai2026}  
\usepackage{times}  
\usepackage{helvet}  
\usepackage{courier}  
\usepackage[hyphens]{url}  
\usepackage{graphicx} 
\usepackage{natbib}  
\usepackage{caption} 
\usepackage{algorithm}
\usepackage{algorithmic}
\usepackage{dsfont}
\usepackage{bbding}
\usepackage{makecell}
\usepackage{booktabs}
\usepackage{multirow}	
\usepackage{amsmath}	
\usepackage{amssymb}	
\usepackage{times}
\usepackage{microtype}
\usepackage{caption}
\usepackage{placeins}
\usepackage{enumitem}
\usepackage{tabularx}
\usepackage{xstring}
\usepackage{xspace}
\usepackage{url}
\usepackage[table]{xcolor}
\usepackage[hang,flushmargin]{footmisc}
\usepackage{soul}
\usepackage{algorithm}
\usepackage{listings}
\usepackage{subfig}

\usepackage{newfloat}
\usepackage{listings}
\DeclareCaptionStyle{ruled}{labelfont=normalfont,labelsep=colon,strut=off} 
\floatstyle{ruled}
\newfloat{listing}{tb}{lst}{}
\floatname{listing}{Listing}
\definecolor{lightblue}{rgb}{0.12,0.49,0.85}

\usepackage[switch]{lineno}
\title{EPSegFZ: Efficient Point Cloud Semantic Segmentation for Few- and Zero-Shot Scenarios with Language Guidance}
\author{
    Jiahui~Wang\textsuperscript{\rm 1}\ Haiyue~Zhu\textsuperscript{\rm 2}\thanks{Corresponding author} \ Haoren~Guo\textsuperscript{\rm 1} \ 
Abdullah~Al~Mamun\textsuperscript{\rm 1}\ 
Cheng~Xiang\textsuperscript{\rm 1} \ 
Tong~Heng~Lee\textsuperscript{\rm 1} \ 
}
\affiliations{
    \textsuperscript{\rm 1}College of Design and Engineering, National University of Singapore\\
    \textsuperscript{\rm 2}SIMTech, Agency for Science, Technology and Research (A*STAR)\\
    \texttt{wjiahui@u.nus.edu}~~~\texttt{zhu\_haiyue@a-star.edu.sg}
}

\begin{document}

\maketitle
\begin{abstract}
Recent approaches for few-shot 3D point cloud semantic segmentation typically require a two-stage learning process, i.e., a pre-training stage followed by a few-shot training stage. While effective, these methods face overreliance on pre-training, which hinders model flexibility and adaptability. Some models tried to avoid pre-training yet failed to capture ample information. In addition, current approaches focus on visual information in the support set and neglect or do not fully exploit other useful data, such as textual annotations. This inadequate utilization of support information impairs the performance of the model and restricts its zero-shot ability. To address these limitations, we present a novel pre-training-free network, named \textbf{E}fficient \textbf{P}oint Cloud Semantic \textbf{Seg}mentation for \textbf{F}ew- and \textbf{Z}ero-shot scenarios. Our EPSegFZ incorporates three key components. A \textbf{Pro}totype-\textbf{E}nhanced \textbf{R}egisters \textbf{A}ttention (\textbf{ProERA}) module and a \textbf{D}ual \textbf{R}elative \textbf{P}ositional \textbf{E}ncoding (\textbf{DRPE})-based cross-attention mechanism for improved feature extraction and accurate query-prototype correspondence construction without pre-training. A \textbf{L}anguage-\textbf{G}uided \textbf{P}rototype \textbf{E}mbedding (\textbf{LGPE}) module that effectively leverages textual information from the support set to improve few-shot performance and enable zero-shot inference. Extensive experiments show that our method outperforms the state-of-the-art method by 5.68\% and 3.82\% on the S3DIS and ScanNet benchmarks, respectively.
\end{abstract}    
\section{Introduction}\label{sec:intro}

\begin{figure}[!t]
    \centering
    \subfloat{
    \includegraphics[width=0.48\linewidth]{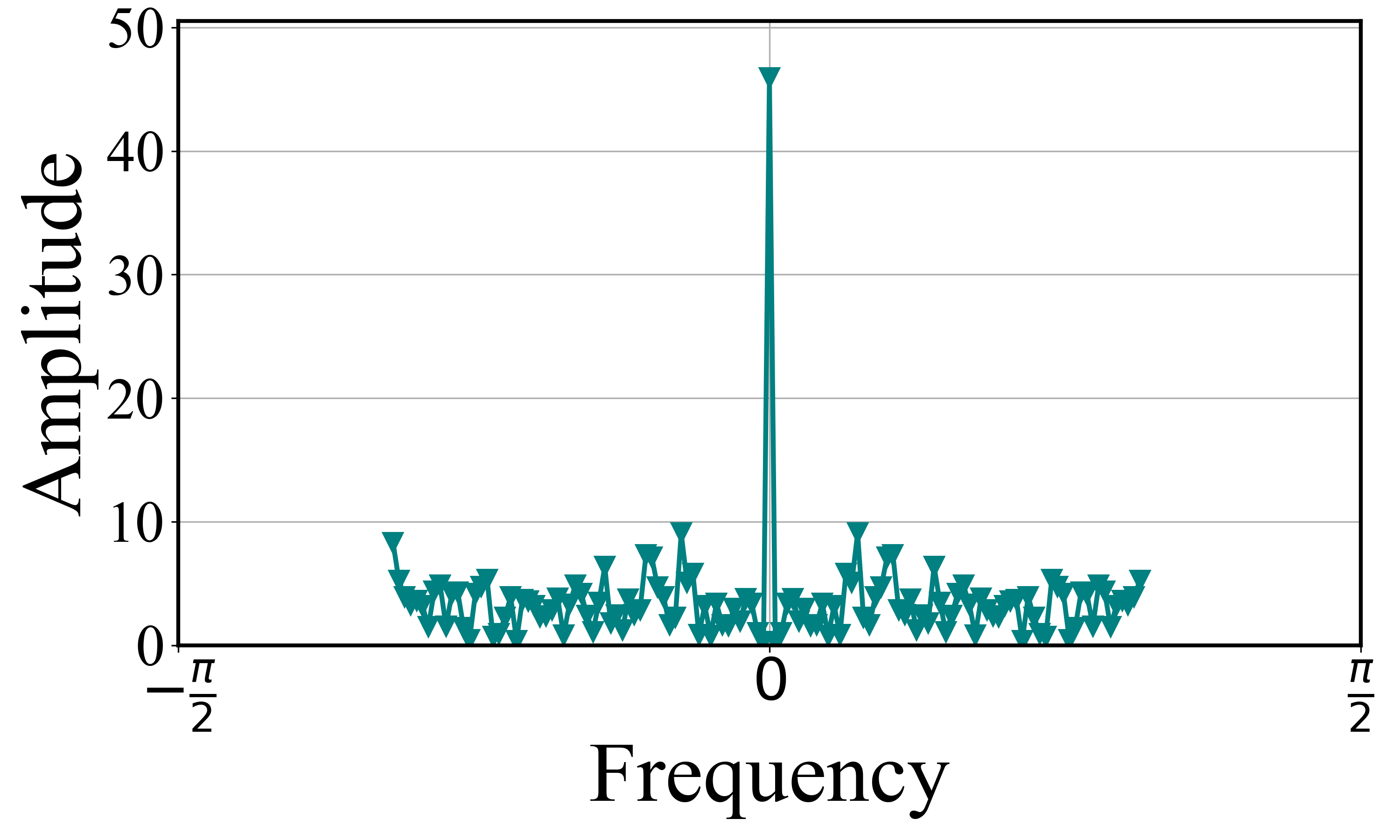}
    }
    \subfloat{
    \includegraphics[width=0.48\linewidth]{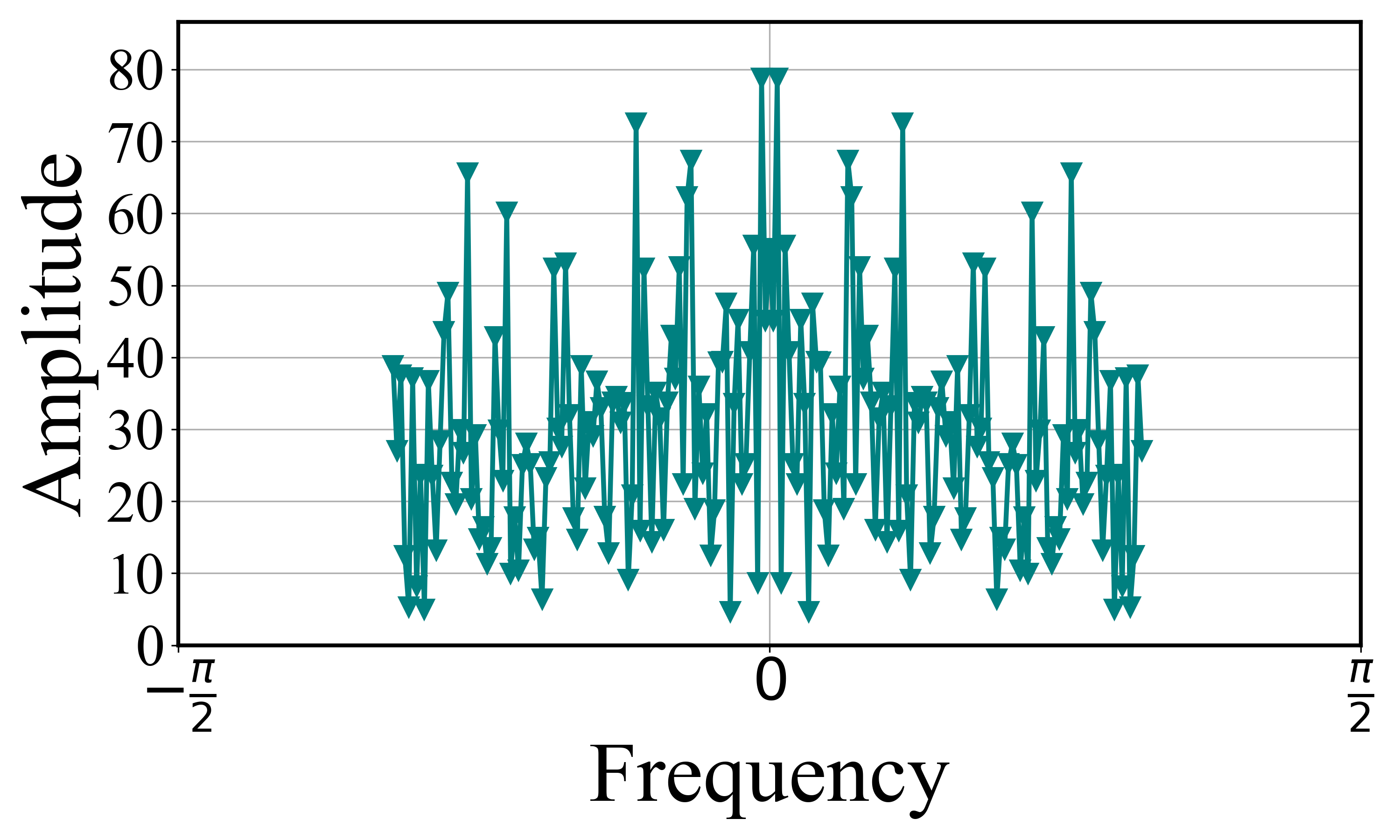}
    }
\caption{Visualized frequency spectrum of embedded features from Seg-PN (\textbf{left}) and Ours (\textbf{right}) (both are pre-training-free method). Our latent features are rich and uniform across frequency bands, while Seg-PN overlooks high-frequency components.}
\label{fig:freq}
\end{figure}

Recently, \textbf{F}ew-\textbf{S}hot \textbf{Sem}antic \textbf{Seg}mentation (\textbf{FS-SemSeg}) for 3D point clouds has gained increasing research interest~\cite {zhao_few-shot_2021,2CBR,lai2022tackling,PAPFZS3D,SCAT}, driven by its potential to efficiently learn from limited data and adapt to unseen categories. However, existing FS-SemSeg approaches heavily rely on fully-supervised pre-trained backbones~\cite{zhao_few-shot_2021,an2024rethinking}, which can introduce biases due to domain differences between datasets. This is particularly problematic as 3D FS-SemSeg datasets are typically small in size and prone to overfitting. Additionally, the pre-training process is resource-intensive and time-consuming, which limits the practical adoption. Seg-PN~\cite{segpn} attempts to address this challenge by designing a non-parametric encoder. However, it discards high-frequency information to ensure robustness (as shown in Figure~\ref{fig:freq}). We argue that despite potential noise, high-frequency features in point clouds carry essential information for precise object segmentation, particularly edge details.
Therefore, developing a method that can both capture high-frequency information effectively and maintain robustness remains a core challenge in FS-SemSeg.

To address this challenge, we developed a trainable attention module called \textbf{Pro}totype-\textbf{E}nhanced \textbf{R}egisters \textbf{A}ttention (\textbf{ProERA}). It utilizes trainable layers to progressively focus on high-frequency information by subtracting low-frequency components and the training process~\cite{xu2019training}. In addition to incorporating register and prototype tokens, ProERA tackles the foreground-background imbalance, placing greater emphasis on high-frequency details typically found in the minority foreground features. Figure~\ref{fig:freq} illustrates the frequency spectrum of prototype features processed by our method compared to Seg-PN~\cite{segpn}. Our prototype features exhibit richness and uniformity across various frequency bands, whereas Seg-PN features predominantly capture low-frequency information.  

While high-frequency information is crucial, achieving a balance between high and low frequencies is essential for a more comprehensive representation. Therefore, low-frequency information should not be overlooked. It is well known that textual embeddings from large pre-trained encoders provide ample low-frequency information \cite{radford2021learning,bai2023qwen,achiam2023gpt}, which can serve as an enhancement in this regard. However, existing models \cite{zhao_few-shot_2021,2CBR,lai2022tackling,SCAT,segpn,an2024rethinking,QGE,wang2025sdsimpoint} primarily rely on point labels and disregard textual information. Given the scarcity of labeled data in FS-SemSeg, leveraging textual cues not only supplements low-frequency information to enhance performance but also maximizes the utility of available data.  

To this end, we propose \textbf{L}anguage-\textbf{G}uided \textbf{P}rototype \textbf{E}mbedding (\textbf{LGPE}), which integrates support text annotations to update prototypes within a unified text-3D joint space. This approach enriches low-frequency representations even during the early training stages and helps the model's learning process. Additionally, LGPE enables prototype construction using only text embeddings, introducing a zero-shot capability.  

Nonetheless, even with features containing rich high and low frequency information, optimal performance cannot be guaranteed. This is because FS-SemSeg prediction relies on query-prototype correspondence, necessitating additional mechanisms to effectively capture these relationships. Previous approaches have attempted to extract such correlations in various ways: COSeg~\cite{an2024rethinking} designed a cascade network that requires numerous training parameters, while Seg-PN~\cite{segpn} calculated Gram matrices for correlation analysis. However, these methods introduce significant computational overhead through additional parameters and substantial memory consumption. 

In response, we propose \textbf{D}ual \textbf{R}elative \textbf{P}ositional \textbf{E}ncoding (\textbf{DRPE}), which introduces \emph{\textbf{no}} extra training parameters. It is the first method to utilize query-prototype relationships within the latent space as Relative Positional Encoding (RPE). Our DRPE process computes the spatial relationships between query and prototype features, incorporating this information as additional input for subsequent cross-attention operations. This approach efficiently captures query-prototype correlations as prior knowledge without resorting to computationally intensive methods. We term this approach DRPE-based cross-attention. The detailed process is illustrated in the Methodology section.

In summary, the main contributions of this work are:
\begin{itemize}
    \item We propose EPSegFZ, a framework that does not require any pre-training and demonstrates the SOTA performance in 3D FS-SemSeg. 
    \item We developed the ProERA module to enhance extracted features with high-frequency, low-noise information, while its sampling strategies and registers mitigate foreground-background imbalance.
    \item An LGPE module is proposed to dynamically update prototypes using support textual data, reducing reliance on perfect support point clouds and enabling zero-shot inference.
    \item A DRPE-based cross-attention is designed as the \textbf{first} to use query-prototype spatial relationships as positional signals, accurately establishing correspondence without additional training burden.

\end{itemize}
\section{Related Work}\label{sec:app_related}
\subsection{Point Cloud Semantic Segmentation} 
Recent research can be generally divided into three main categories. The voxel-based, superpoint-based, and point-based approaches. As documented in~\cite{meng2019vv,choy20194d,vox1,vox2,vox3}, partition the 3D space into regular voxels or superpoints before employing sparse convolutions on them. While these methods demonstrate reasonable performance, they are hindered by imprecise position data resulting from the partition process. The point-based methods~\cite{qi_pointnet_2017,thomas_kpconv_2019,wu2024point,he2024segpoint,zhang2024pointgt} directly consider the features and position of each point as inputs, thus preserving natural information. KPConv~\cite{thomas_kpconv_2019} utilized carefully designed convolution kernels to capture multi-level information. PointNet++~\cite{qi_pointnet_2017} trains its Multi-Layer Perceptron (MLP) with point clouds sampled by different strategies to capture global and local information. PointTransformer~\cite{wu2024point} utilizes the attention mechanism to learn point-wise information effectively.
\subsection{Few-Shot Point Cloud Semantic Segmentation}
AttMPTI~\cite{zhao_few-shot_2021}, a groundbreaking approach, utilizes label propagation to uncover connections between prototypes and query points with features extracted by a pre-trained backbone. 2CBR~\cite{2CBR} harnesses shared features of support and query to compute bias factors and correct disparities between them. PEFC~\cite{zhang2023PEFC} implements two specialized components to expand the prototype set and adjust them using query characteristics in a two-way, context-aware fashion. PAPFZS3D~\cite{PAPFZS3D} presents a prototype adaptation framework that simultaneously enhances both prototypes and query features. SCAT~\cite{SCAT} introduces a layered, class-specific attention-based transformer, establishing detailed associations between support and query features. COSeg~\cite{an2024rethinking} introduced a correlation memory-based network to reveal the inherent relationship between queries and prototypes. Seg-PN~\cite{segpn} proposes a computationally efficient model employing a non-parametric encoder and correlation-driven support-query interactions to determine point-wise classifications. MM-FSS~\cite{an2024multimodalityhelpsfewshot3d} uses a multi-modality framework to fuse 2D and 3D information for more details. However, the significant usage of computational resources limited its application in lightweight scenarios.
\begin{figure*}[t]
\centering
\includegraphics[width=0.92\linewidth]{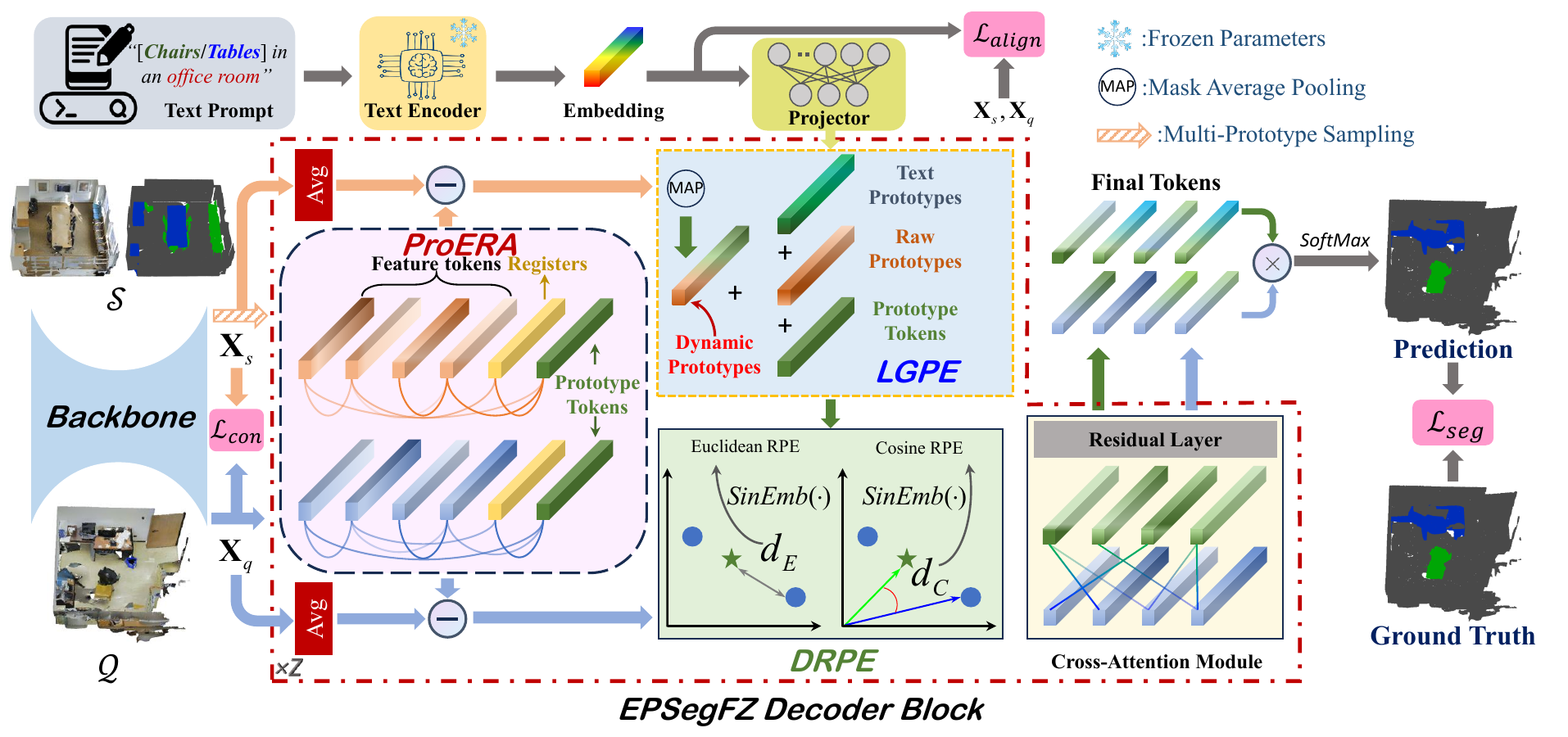}
\caption{The visualized architecture of our EPSegFZ. A ProERA module first captures high-frequency information and refines the extracted feature. Then, an LGPE module dynamically updates the class prototypes with textual embeddings. After that, a DRPE-based cross-attention properly builds correspondence between prototypes and query features. Finally, the prediction result is obtained by dot production. The red block Avg. represents the average pooling operation.
}
\label{fig:architecture}
\end{figure*}
\begin{figure}[t]
    \centering
    \makebox[\linewidth]{\subfloat[PAPFZS3D]{
    \includegraphics[width=0.3\linewidth]{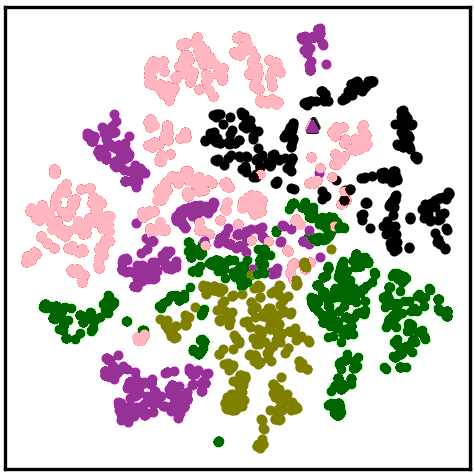}
    }
    \hfill
    \subfloat[Seg-PN]{\includegraphics[width=0.3\linewidth]{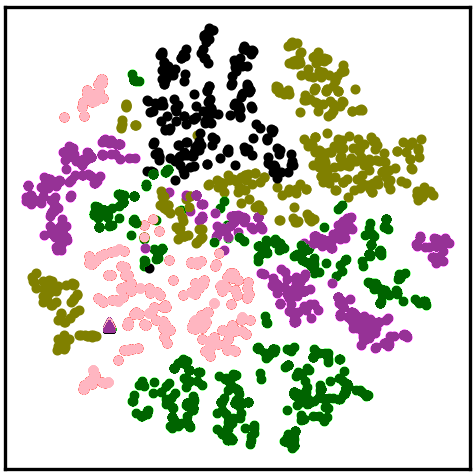}}
    \hfill
    \subfloat[Ours]{
    \includegraphics[width=0.3\linewidth]{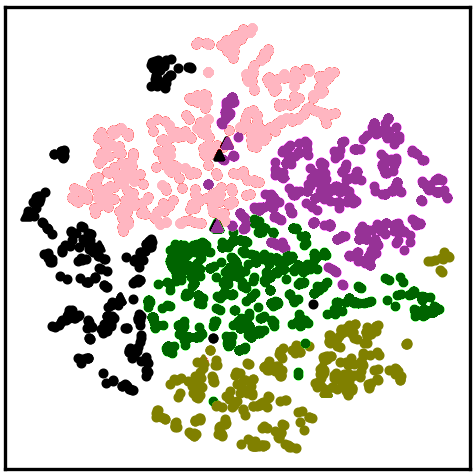}
    }}
\caption{Visualized t-SNE embedding of feature tokens for prediction. With our LGPE and DRPE, same-class features form a more compact distribution, enhancing the discriminative ability. Colored points represent semantic classes.
}
\label{fig:tsne}
\end{figure}

\section{Methodology}

\subsection{Preliminary}

The FS-SemSeg task to be addressed in this work is an $N$-way $K$-shot problem~\cite{zhao_few-shot_2021,segpn}. Consider a support set $\mathcal{S}=\{(\mathbf{P}_{s}^{n,k}, \mathbf{Y}_{s}^{n,k})_{k=1}^{K}\}_{n=1}^{N}$, where $N$ is the number of classes, each class consists of $K$ support point cloud samples $\mathbf{P}_{s}$ with the corresponding labels $\mathbf{Y}_{s}$. Given the query set $\mathcal{Q}=(\mathbf{P}_{q}, \mathbf{Y}_{q})$, $\mathbf{Y}_{q}$ is the ground truth which is not available for inference, the goal of this work is to obtain a desirable network $\mathcal{N}_{\theta}(\cdot)$ and its parameter $\theta$ with the objective:
\begin{equation}
\underset{\theta}{\arg \max } \prod_{\mathcal{Q}} p\left(\mathbf{Y}_{q} \mid \mathcal{N}_{\theta}\left(\mathbf{P}_{q} \mid \mathcal{S}\right)\right).
\end{equation}

\subsection{Overview}

The overall training architecture of our proposed method is illustrated in Figure~\ref{fig:architecture}. For each episode in the training, point clouds from the support and query sets are mapped into a latent space using a DGCNN (train from scratch)~\cite{wang_dynamic_2019}. The Multi-Prototype Sampling (MPS) technique~\cite{zhao_few-shot_2021} is then applied to the support features to derive prototypes. To refine these features and capture high-frequency details, the ProERA module appends learnable registers and prototype tokens to mitigate noise and facilitate query-prototype interaction. Notice that, in each block of the ProERA module, we subtract the average of the input features from the attention output, yielding high-frequency-dominant features. Then, LGPE updates the prototypes with the textual embedding from a language model and adds this to the dynamic prototype obtained by mask average pooling. As shown in Figure~\ref{fig:tsne}, t-SNE visualizations compare baselines with our EPSegFZ, demonstrating that our approach achieves a more refined representation space. Next, the DRPE module calculates the correlation between the refined query features and updated prototypes, encoding this information into the subsequent cross-attention module~\cite{qin2022geometric} for precise correspondence establishment. Finally, we formulate the prediction by computing the similarity based on the dot product between prototypes and query points~\cite{Strudel_2021_segmenter, PAPFZS3D}. 

\subsection{Prototype-Enhenced Registers Attention}\label{sec:ProERA}
The attention mechanism has been proven effective in extracting rich semantic and contextual features~\cite{zhao_few-shot_2021,dosovitskiy2021vit}. However, applying attention to all support points can overwhelm the model with irrelevant background data, thus impairing the prototype quality. Besides, early in training, networks prioritize low-frequency information from backgrounds~\cite{xu2019training}, while our target objects (foreground), represented by fewer points, contain higher-frequency data due to their complex geometry and semantics that might be overlooked.

\begin{figure}[t]
    \centering
    \includegraphics[width=1.0\linewidth]{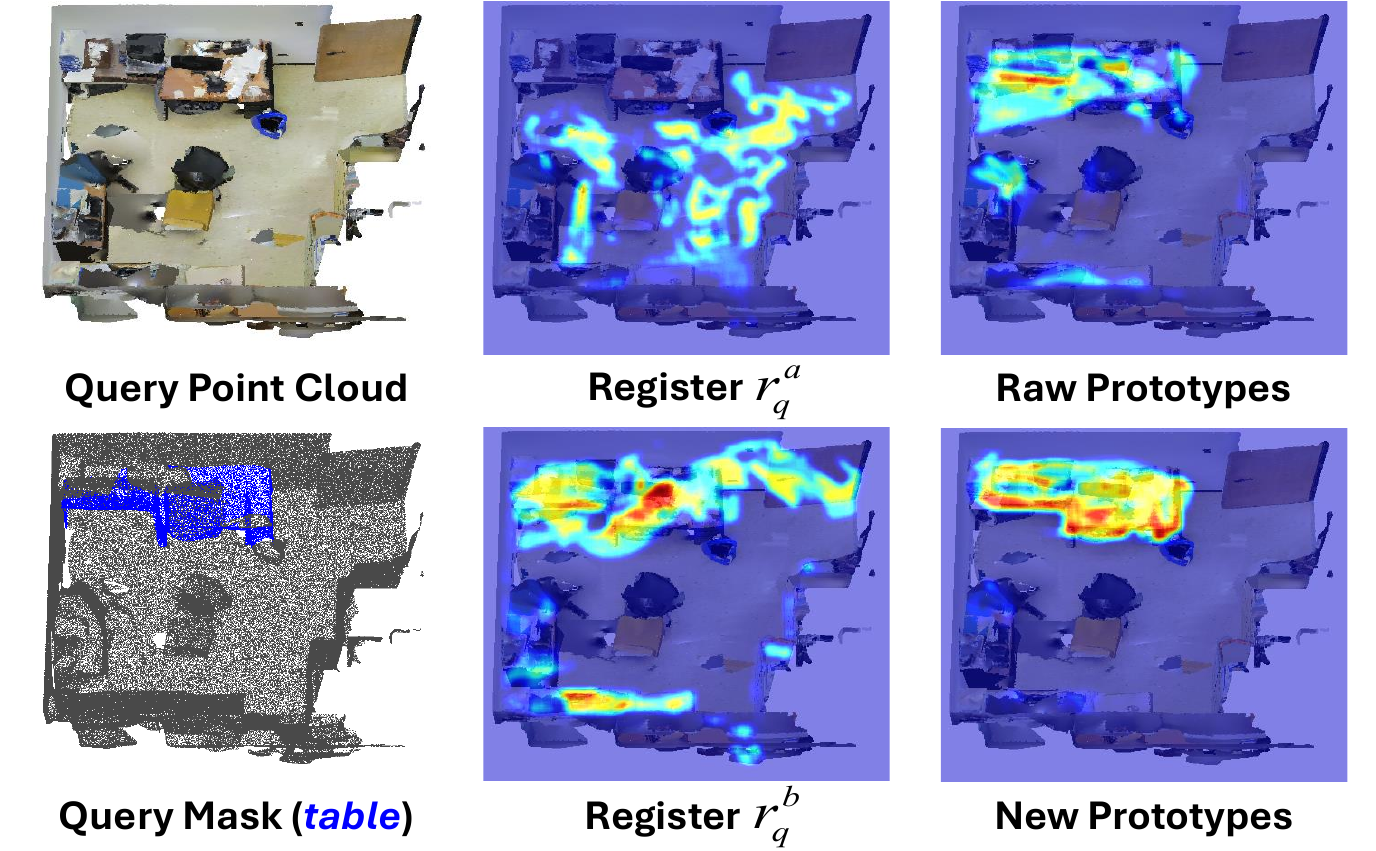}
    \caption{Visualized heatmaps of query-registers and query-prototypes similarities. The distinct focused region of registers helps the model differentiate between object-related and object-less areas. The updated prototypes effectively correlate with the query object, whereas the raw prototypes lack sufficient focus on the target object.
    }
    \label{fig:simmap}
\end{figure}

Considering this, we replace support points with multi-prototypes sampled from support data and add registers for both query and prototype tokens. This approach reduces computational load and mitigates the potential influence of a large number of irrelevant background points. Moreover, self-attention usually acts as a low-pass filter~\cite{wang2022antioversmoothing}, emphasizing low-frequency over high-frequency. Therefore, we use the average pooling operation to pre-calculate the overall low-frequency information and subtract such a feature from the output to produce features containing sufficient high-frequency information. Figure~\ref{fig:simmap} illustrates the similarity map between query features and registers. It suggests that the registers prioritize distinct areas within the scene: one register directs attention toward the background and object-free area, whereas the other emphasizes the area that notably contains multiple objectives. This method also mitigates the background-foreground imbalance, which implicitly addresses low-frequency vs. high-frequency imbalance during training, thereby enhancing the representational capability.

Specifically, given a support point cloud $\mathbf{P}_{s}$ and a query point cloud $\mathbf{P}_{q}$, each consisting of $M$ points, their features are first obtained with a backbone and denoted as $\mathbf{X}_t \in \mathbb{R}^{M\times D}; t\in\{s,q\}$ where $D$ is the embedded dimension. 
In our ProERA module, we append extra $n_r$ learnable tokens, denoted as $\mathbf{r}_{t} \in \mathbb{R}^{n_r \times D}$, as shown in the Figure~\ref{fig:architecture}. 
Besides, to strengthen the interactions between prototypes and these features, we additionally append prototype tokens behind registers. The multi-prototype $\mathbf{X}_{p}$ and raw prototype token $\mathbf{p}_{raw}$ of the $c$-th category are denoted as:
\begin{equation}
   \mathbf{X}_p= \text{MPS}\left(\mathbf{X}_s\right); \mathbf{p}_{raw}^c=\frac{1}{n_p} \sum_{n_p}(\mathbf{X}_p \times \mathds{1}(\mathbf{Y}_p = c)),
\end{equation}
where $n_p$ is a hyperparameter determining the number of sampled prototypes for each class, $\text{MPS}(\cdot)$ is the multi-prototype sampling algorithm~\cite{zhao_few-shot_2021}, and $\mathbf{Y}_p$ is the label of the prototype. The function $\mathds{1(\cdot)}$ is the binary label indicator that outputs 1 when its variable is true.
Consider the $i$-th decoder block ($i \in [1, Z]$), with the residual layer $\text{Res}(\cdot)$ and self-attention layer $\text{SA}(\cdot)$ the output of our ProERA module is:
\begin{equation}
    \tilde{\mathbf{X}}_{j}^{i}= \text{Res}\left(\text{SA}([\hat{\mathbf{X}}_{j}^{i-1};\mathbf{r}_{j};\hat{\mathbf{p}}^{i-1}])\right) - \frac{1}{n_j}\sum_{n_j}\hat{\mathbf{X}}_j^{i-1},
\end{equation}
where $[;;]$ is the concatenation operation, $j\in\{p,q\}$ denotes equal operation to query and support. $\hat{\mathbf{p}}^{i-1}$ and $\hat{\mathbf{X}}_{j}^{i-1} $ denote the prototype tokens and feature tokens from the last block, respectively; their definition will be illustrated in the following section. 
For the first decoder block, the relative inputs are $\hat{\mathbf{X}}_{j}^{0} =\mathbf{X}_{j}$ and $\hat{\mathbf{p}}^{0}=\mathbf{p}_{raw}$. 

\subsection{Language-Guided Prototype Embedding}\label{sec:LGPE}
Due to the limited data available in FS-SemSeg, the declined representation of prototypes caused by imperfect support point clouds hinders the learning process. Moreover, if the backbone is not pre-trained, the prototypes in the early stage of training are not discriminative. These findings prompt us to reduce the model's reliance on solely visual support information.  We propose an LGPE module to optimize the prototypes and enable the network to conduct zero-shot inference using support text labels.

Our LGPE module receives the embedding feature from the text encoder of a pre-trained CLIP~\cite{radford2021learning} that is prompted with corresponding support classes to form text prototypes:
\begin{equation}
        \mathbf{p}_{text}^c = \text{Proj}\left(\mathbf{T}^c\right),
\end{equation}
where $\mathbf{T}^c$ refers to a text embedding from the CLIP text encoder for the $c$-th class, and $\text{Proj}(\cdot)$ is a projection network. Take the $i$-th block in the decoder as an example, the dynamic prototypes are obtained with the current multi-prototype $\Tilde{\mathbf{X}}_p^i$ and mask average pooling, i.e.,
$
    \mathbf{p}_{dyn}^{i,c} = \frac{1}{n_p} \sum_{n_p} \tilde{\mathbf{X}}_p^{i,c}.
$

Subsequently, the updated prototypes represent a blend of previous prototype tokens, the raw prototypes, the text prototypes, and the dynamic prototypes, as indicated by:
\begin{equation}
    \mathbf{p}^{i} = \lambda_1 \Tilde{\mathbf{p}}^{i} +\lambda_2 \mathbf{p}_{raw}+\lambda_3 \mathbf{p}_{dyn}^i + \lambda_4\mathbf{p}_{text},
\end{equation}
where $\Tilde{\mathbf{p}}^{i}$ is obtained from $\Tilde{\mathbf{X}}_p^{i}$ with mask average pooling. Considering that text embedding is already well-established while visual embedding starts with a weaker representation, their fusion into the prototype should dynamically adjust over time. Initially, the model relies more on the text embedding to provide strong guidance, as the visual embedding has not yet learned meaningful features. As training progresses and the visual representation improves, its weight gradually increases, allowing the model to shift towards a more balanced fusion of both modalities. This mechanism ensures a smooth transition from text-driven supervision to a learned visual-text alignment.  Formally, we define the weight of the text embedding as \(\lambda_4(t) = \lambda_4^{*}e^{-0.5t}\), where \(\lambda_4^{*}\) is a predefined final weight. Meanwhile, the weights for visual embeddings, \(\lambda_i(t), t \in [1,3]\), follow an increasing function \(\lambda_i(t) = \lambda_i^{*}(1 - e^{-0.5t})\), ensuring a gradual adaptation that shifts the prototype’s reliance from text to learned visual features over time. This processing not only provides favorable prototypes but also mitigates the influence of the random-initialized backbone in the early stage of training. Figure~\ref{fig:simmap} illustrates that employing the updated prototypes contributes to establishing preferable correspondence, thereby facilitating more precise predictions.

\subsection{Dual Relative Positional Encoding}\label{sec:DRPE}
Our DRPE module is the first to incorporate the inter-cloud correlation (query-prototype) in the latent space as a positional encoding signal in cross-attention for FS-SemSeg tasks. Specifically, the Euclidean distance between the $j$-th point in the query point cloud and the $c$-th prototype in the $i$-th block is $d_{E}^{i,j,c}$.
Hence, the distance between all query points and the $c$-th prototype is a vector that can be expressed as $\mathbf{d}_E^{i,c} \in \mathbb{R}^{M \times 1}$. This vector is then used as indices to calculate the encoding value $\mathbf{R}_{E}^{i,c} \in \mathbb{R}^{M \times(N+1) \times D}$
with the sinusoidal positional encoding function~\cite{vaswani2017attention}. 
Simultaneously, the cosine value of the angle formed between a query vector and a prototype vector is calculated and indicated as $d_{C}^{i,j,c}$.
This distance is also encoded with the same sinusoidal function and denoted as $\mathbf{R}_{C}^{i} \in \mathbb{R}^{M \times(N+1)\times D}$. The DRPE value is an element-wise addition of the two results,  $\mathbf{R}^{i} = \mathbf{R}_{C}^{i} + \mathbf{R}_{E}^{i}.$,
With the calculated DRPE, we can formulate a DRPE-based decoder; its visualized architecture is shown in the Appendix. 

For prediction, we follow~\cite{Strudel_2021_segmenter} to conduct the pipeline. The query feature $\Tilde{\mathbf{X}}_q^{i}$ and the updated prototypes $\mathbf{p}^i$ are fed into a cross-attention module, then processed sequentially.
The final query feature token $\hat{\mathbf{X}}_q^Z$ and prototypes $\hat{\mathbf{p}}^Z$ are the output of the $Z$-th decoder which are used to generate the prediction result
\begin{equation}
\hat{\mathbf{Y}}_q=\text{SoftMax}(\frac{\hat{\mathbf{X}}_q^Z}{||\hat{\mathbf{X}}_q^Z||_2} \cdot \frac{\hat{\mathbf{p}}^Z}{||\hat{\mathbf{p}}^Z||_2}).
\end{equation}
Each row in $\hat{\mathbf{Y}}_q$ represents the probability that the corresponding point belongs to each support class. Detailed calculation process within the cross-attention can be accessed in the appendix.
\subsection{Loss Functions}\label{sec:loss}
To achieve better performance, the loss function must be carefully designed. 
Since our backbone is not pre-trained, we propose a foreground-consistency loss to enhance its feature representation, as shown in Equation~\ref{eq:lcons}. It encourages positive pairs, features from the same foreground region, to be closer in the embedding space while pushing negative pairs apart. By focusing on foreground consistency, our method helps the backbone learn more discriminative features while mitigating instability caused by the lack of pre-training. Thus, it is defined as:
\begin{equation}
    \mathcal{L}_{con}= \text{InfoNCE}(\mathbf{x}_q,\mathbf{x}_s)
    \label{eq:lcons}
\end{equation}
where $\text{InfoNCE}(\cdot)$ is an InfoNCE-based~\cite{he_momentum_2020} constrastive loss, $\tau$ is a temperature hyperparameter. The detailed formulation of $\mathcal{L}_{con}$ is illustrated in the appendix.

Building on this foundation, we introduce a foreground-aware alignment loss that serves two key purposes: reducing the gap between visual and text embeddings to enhance the text-visual joint space while also providing crucial guidance during early training stages. This loss function minimizes the cross-entropy between text-visual similarity and text labels, encouraging better alignment across modalities. It ensures that the features of the foreground objects closely match their textual descriptions, leading to a more meaningful relationship between image and text and improved performance in multimodal tasks. Mathematically, the loss can be written as:
\begin{equation}
    \mathcal{L}_{align} = \frac{1}{N}\sum_{c=1}^N \text{CE}\left(\text{SoftMax}(W\mathbf{p}_{raw}^c \cdot \mathbf{T}^{c}), c\right),
\end{equation}
where $W$ is a learnable matrix, $\text{CE}(\cdot)$ is the cross-entropy loss. $\mathbf{p}_{raw}^c$ is the visual prototype for the $c$-th foreground, and $\mathbf{p}_{text}$ is a feature map that represents text embeddings of all foreground objects. 

Lastly, the main segmentation loss supervises all modules by minimizing the cross-entropy between predicted $\hat{\mathbf{Y}}_{q}$ and the ground-truth $\mathbf{Y}_q$: $\mathcal{L}_{seg}= \text{CE}(\mathbf{Y}_{q}, \hat{\mathbf{Y}}_{q}).$
Therefore, our final loss in training is:
$\mathcal{L}=\mathcal{L}_{seg}+\lambda_{con}\mathcal{L}_{con}+\lambda_{align}\mathcal{L}_{align}$

\section{Experiments}
\begin{table}[t]
\centering
    \resizebox{0.47\textwidth}{!}{\begin{tabular}{c c c c c}
    \toprule
     Model & $\#$Params.& GFLOPs& Time(s) &$\Delta$ (m-IoU)\\
    \midrule
    AttMPTI &372.19K &2.60 &0.46 &{-19.31}\\
    PAPFZS3D &2.46M	&3.07 &0.62 &{-13.63}\\
    Seg-PN
    &241.67K	&1.95 &0.32 &{-8.24}\\
     COSeg &7.69M	&9.71	&1.35 & +0.05\\
    \rowcolor{lightblue!10}
    \textbf{Ours} &2.02M	&2.11	&0.36 & - \\
    \bottomrule
    \end{tabular}}
    \caption{Efficiency analysis on S3DIS based on model complexity, FLOPs, and inference time.}
    \label{tb:modelcomp}
\end{table}
\begin{table*}[ht]
    \centering
    \resizebox{1.0\linewidth}{!}{\begin{tabular}{l c c c c c c c c c c c c c c c c}
    \toprule 
    \multirow{3}{*}{ Method }& \multicolumn{4}{c}{2-way 1-shot} & \multicolumn{4}{c}{2-way 5-shot} & \multicolumn{4}{c}{3-way 1-shot} & \multicolumn{4}{c}{3-way 5-shot} \\
    \cmidrule(lr){2-5}  \cmidrule(lr){6-9} \cmidrule(lr){10-13} \cmidrule(lr){14-17} 
    & $S^0$ & $S^1$ & mean &$\Delta$ & $S^0$ & $S^1$ &  \multicolumn{1}{c}{mean} &$\Delta$  & \multicolumn{1}{c}{$S^0$} & $S^1$ & mean &$\Delta$  & $S^0$ & $S^1$ & mean &$\Delta$ \\
    \midrule 
    Fine-Tuning & 36.34 & 38.79 & 37.57 &\textcolor{black}{-35.85} & 56.49 & 56.99 & 56.74 &\textcolor{black}{-18.27} & 30.05 & 32.19 & 31.12&\textcolor{black}{-34.81} & 46.88 & 47.57 & 47.23&\textcolor{black}{-21.05} \\
    ProtoNet & 48.39 & 49.98 & 49.19 &\textcolor{black}{-25.89} & 57.34 & 63.22 & 60.28&\textcolor{black}{-15.73}
& 40.81 & 45.07 & 42.94&\textcolor{black}{-22.99}
& 49.05 & 53.42 & 51.24&\textcolor{black}{-17.04}
\\
    AttMPTI & 53.77 & 55.94 & 54.86&\textcolor{black}{-18.56}
& 61.67 & 67.02 & 64.35 &\textcolor{black}{-11.66}
& 45.18 & 49.27 & 47.23&\textcolor{black}{-18.70}
& 54.92 & 56.79 & 55.86&\textcolor{black}{-12.42}
\\
    CWT &52.14 &57.86 &55.00&\textcolor{black}{-18.42}
&61.64 &66.48 &64.06&\textcolor{black}{-11.95}
&- &- &-&-
&- &- &-&-
\\
    2CBR &55.89 &61.99 &58.94&\textcolor{black}{-14.48}
&63.55 &67.51 &65.53&\textcolor{black}{-10.48}
&46.51 &53.91 &50.21&\textcolor{black}{-15.72}
&55.51 &58.07 &56.79&\textcolor{black}{-11.49}
\\
    SCAT &54.92 &56.74 &55.83&\textcolor{black}{-17.59}
&64.24 &69.03 &66.63&\textcolor{black}{-9.38}
&- &- &-&-
&- &- &-&-
\\ 
    PEFC &55.09 &59.63 &57.36&\textcolor{black}{-16.06}
&65.47 &70.84 &68.16&\textcolor{black}{-7.85}
&49.15 &54.69 &51.92&\textcolor{black}{-14.01}
&62.56 &63.21 &62.89&\textcolor{black}{-5.39}
\\
    QGE  &58.85 &60.29& 59.57&\textcolor{black}{-13.85}
&66.56 &\textbf{79.46} &69.01&\textcolor{black}{-7.00}
&-&-&-&-
&-&-&-&-
\\
    PAPFZS3D &59.45 &66.08 &62.76&\textcolor{black}{-10.66}
&65.40 &70.30 &67.85 &\textcolor{black}{-8.16}
&48.99 &56.57 &52.78&\textcolor{black}{-13.15}
&61.27 &60.81 &61.04&\textcolor{black}{-7.24}
\\
    Seg-PN &64.84 &67.98 &66.41&\textcolor{black}{-7.01}&67.63 &71.48 &69.36&\textcolor{black}{-6.65}&60.12 &63.22 &61.67&\textcolor{black}{-4.26}&62.58 &64.53 &63.56&\textcolor{black}{-4.72}\\

SDSimPoint &{68.73} &{70.61} &{69.67}&\textcolor{black}{-3.75} &{72.12} &{72.72} &{72.42}&\textcolor{black}{-3.59} &{62.28} &{62.11} &{62.19}&\textcolor{black}{-3.74} &{65.17}	&{66.10} &{65.64}&\textcolor{black}{-2.64}\\
    
    \rowcolor{lightblue!10}\textbf{Ours+Point-NN} &\underline{72.31} &\textbf{74.20} &\underline{73.26}&\textcolor{black}{-0.16} &\underline{75.26} &{75.94} &\underline{75.60} &\textcolor{black}{-0.41}&\textbf{65.63}&\underline{66.09}&\underline{65.86}&\textcolor{black}{-0.07}&\underline{67.94} &\textbf{68.55} &\underline{68.25}&\textcolor{black}{-0.03}\\
    \rowcolor{lightblue!10}
    \textbf{Ours+DGCNN} &\textbf{73.08} &\underline{73.75} &\textbf{73.42} &-&\textbf{75.90}&\underline{76.11}&\textbf{76.01}&-&\underline{65.58}&\textbf{66.27}&\textbf{65.93}&-&\textbf{68.30} &\underline{68.25} &\textbf{68.28}&-\\
    \toprule
    \end{tabular}}
     \caption{Evaluation result on the S3DIS dataset using mean-IoU criteria (\%). The best result of each column is highlighted with \textbf{bold font}, and the second best is noted with \underline{underline}.}
    \label{tb:s3dis}
\end{table*}
\begin{table*}[t]
    \centering
    
    \resizebox{1\textwidth}{!}{\begin{tabular}{l c c c c c c c c c c c c c c c c}
    \toprule
    \multirow{3}{*}{ Method } & \multicolumn{4}{c}{2-way 1-shot} & \multicolumn{4}{c}{2-way 5-shot} & \multicolumn{4}{c}{3-way 1-shot} & \multicolumn{4}{c}{3-way 5-shot} \\
    \cmidrule(lr){2-5}  \cmidrule(lr){6-9} \cmidrule(lr){10-13} \cmidrule(lr){14-17} 
    & $S^0$ & $S^1$ & mean &$\Delta$ & $S^0$ & $S^1$ &  \multicolumn{1}{c}{mean} &$\Delta$  & \multicolumn{1}{c}{$S^0$} & $S^1$ & mean &$\Delta$  & $S^0$ & $S^1$ & mean &$\Delta$ \\
    \midrule 
    Fine-Tuning & 31.55 & 28.94 & 30.25 &\textcolor{black}{-38.59}
& 42.71 & 37.24 & 39.98 &\textcolor{black}{-30.66}
& 23.99 & 19.10 & 21.55&\textcolor{black}{-45.48}
& 34.93 & 28.10 & 31.52&\textcolor{black}{-38.21}
\\
    ProtoNet & 33.92 & 30.95 & 32.44&\textcolor{black}{-36.40}
& 45.34 & 42.01 & 43.68&\textcolor{black}{-26.96}
& 28.47 & 26.13 & 27.30&\textcolor{black}{-39.73}
& 37.36 & 34.98 & 36.17&\textcolor{black}{-33.56}
\\
    AttMPTI & 42.55 & 40.83 & 41.69&\textcolor{black}{-27.15}
& 54.00 & 50.32 & 52.16&\textcolor{black}{-18.48}
& 35.23 & 30.72 & 32.98&\textcolor{black}{-34.05}
& 46.74 & 40.80 & 43.77&\textcolor{black}{-25.96}
\\
    CWT &42.33 &41.78 &42.05&\textcolor{black}{-26.79}
&55.60 &53.77 &56.48&\textcolor{black}{-14.16}
&- &- &-&-
&- &- &-&-
\\
    2CBR &50.73 &47.66 &49.20&\textcolor{black}{-19.64}
&52.35 &47.14 &49.75&\textcolor{black}{-20.89}
&47.00 &46.36 &46.68&\textcolor{black}{-20.35}
&45.06 &39.47 &42.27&\textcolor{black}{-27.46}
\\
    SCAT  &45.24 &45.90 &45.57&\textcolor{black}{-23.27}
&55.38 &57.11 &56.24&\textcolor{black}{-14.40}
&- &- &- &-
&- &-&-&-
\\ 
    PEFC &45.31 &44.86 &45.09&\textcolor{black}{-23.75}
&56.26 &54.06 &55.16 &\textcolor{black}{-15.48}
&38.78 &36.13 &37.46&\textcolor{black}{-29.57}
&51.72 &46.05 &48.89&\textcolor{black}{-20.84}
\\
    
     QGE  &43.10 &46.79 &44.95&\textcolor{black}{-23.89}
&51.91 &57.21 &54.56&\textcolor{black}{-16.08}
&-&- &-&-
&-&-&-&-
\\
    PAPFZS3D &57.08 &55.94 &56.51&\textcolor{black}{-12.33}
&64.55 &59.64 &62.10 &\textcolor{black}{-8.54}
&55.27 &55.60 &55.44&\textcolor{black}{-11.59}
&59.02 &53.16 &56.09&\textcolor{black}{-13.64}
\\
    Seg-PN &63.15 &64.32 &63.74&\textcolor{black}{-5.10}
&67.08 &69.05 &68.07&\textcolor{black}{-2.57}
&61.80 &65.34 &63.57 &\textcolor{black}{-3.46}
&62.94 &68.26 &65.60&\textcolor{black}{-4.13}\\

SDSimPoint &{65.21} &{65.18} &{65.19}&\textcolor{black}{-3.65} &{68.20} &68.49 &{68.35}&\textcolor{black}{-2.29} &{63.30} &63.86 &{63.83}&\textcolor{black}{-3.20} &{65.04} &66.27 &{65.66}&\textcolor{black}{-4.07}

\\
    \rowcolor{lightblue!10}\textbf{Ours+Point-NN} &\underline{65.71} &\underline{66.01} &\underline{65.86}&\textcolor{black}{-2.98}&\underline{67.94} &\underline{69.22} &\underline{68.58}&\textcolor{black}{-2.06}&\underline{64.32} &\underline{65.57} &\underline{64.95}&\textcolor{black}{-2.08}&\underline{66.32}	&\underline{68.74} &\underline{67.53}&\textcolor{black}{-2.20}\\
    \rowcolor{lightblue!10}
    \textbf{Ours+DGCNN} &\textbf{69.43} &\textbf{68.25} &\textbf{68.84}&- &\textbf{71.31} &\textbf{69.97} &\textbf{70.64} &-&\textbf{67.19} &\textbf{66.86} &\textbf{67.03}&- &\textbf{69.62} &\textbf{69.83} &\textbf{69.73}&-\\
    \toprule
    \end{tabular}}
    \caption{Evaluation result on the ScanNet dataset using mean-IoU criteria (\%). The best result of each column is highlighted with \textbf{bold font}, and the second best is noted with \underline{underline}.}
    \label{tb:scannet}
    \vspace{-5mm}
\end{table*}
\subsection{Implementation Details}
\textbf{Data preparation.} We construct our FS-SemSeg tasks on the S3DIS~\cite{s3dis} and the ScanNet~\cite{scannet} datasets. As in previous works~\cite{zhao_few-shot_2021,he2023prototype,wang2023few-shotICRA}, we adopt the preprocessing methodology proposed in~\cite{qi_pointnet_2017}, which involves splitting rooms into smaller units.
For training, we randomly selected 2048 points per unit. We partition each dataset into two distinct subsets, $S^0$ and $S^1$, as prior methodologies~\cite{zhao_few-shot_2021,he2023prototype}, where one subset exclusively serves for inference, and the other is designated for training. 

\noindent\textbf{Training and Inference.} As in previous works~\cite{he2023prototype,zhang2023PEFC}, episodic learning is adopted in the training and testing. Our total iteration number is set to 30,000. To obtain a meaningful visual representation in the early training stage, we set a relatively large learning rate for the backbone to enable faster updates. We also assign a smaller decay step, which decreases more rapidly than the learning rate of other modules. Moreover, we implement a random shuffle of the point order in the query and support the data for fairness. During inference, 100 episodes are randomly selected. Text embeddings are obtained in advance and saved locally as a dictionary to mitigate computational burden. For detailed hyperparameter settings, please refer to the appendix. 
\begin{figure}[!h]
    \centering
    \includegraphics[width=0.95\linewidth]{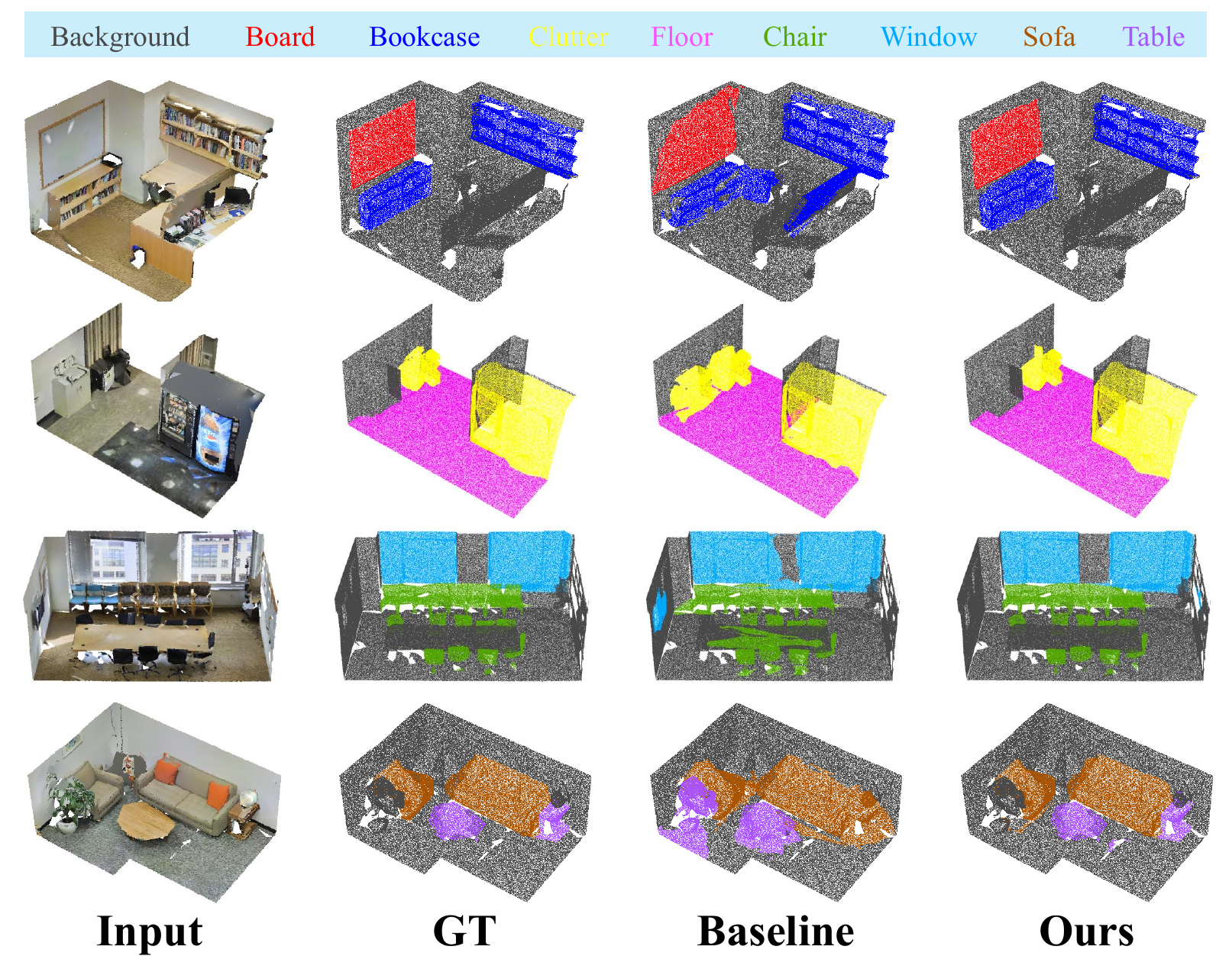}
\caption{Visualized segmentation result on S3DIS dataset. Our method performs better in segmentation accuracy than the baseline.}
\label{fig:visres}
\end{figure}
\subsection{Experimental Results}\label{sec:expresult}
Table~\ref{tb:modelcomp}  shows the efficiency analysis ($\Delta$ represents the gap between ours and the corresponding method). Our method demonstrates efficiency with comparable FLOPs and inference time, given the enhanced accuracy, making it a competitive approach. Table~\ref{tb:s3dis} and Table~\ref{tb:scannet} present the performance of the different models in the S3DIS~\cite{s3dis} and ScanNet~\cite{scannet} datasets. We tested our approach with the trainable DGCNN backbone~\cite{wang_dynamic_2019} and the nonparametric Point-NN backbone~\cite{zhang2023parameter} used in Seg-PN~\cite{segpn}. Our proposed approach demonstrates a substantial improvement in all FS-SemSeg settings and outperforms previous SOTA by \textbf{5.68\%} and \textbf{3.82\%} m-IoU on S3DIS~\cite{s3dis} and ScanNet~\cite{scannet}, respectively. We introduce each compared baselines in the appendix. Figure~\ref{fig:visres} depicts the visualized segmentation results of our method and baseline (Seg-PN~\cite{segpn}); in contrast to the baseline, our EPSegFZ exhibits clearer segmentation on edges, benefiting from ample high-frequency information. 

\subsection{Zero-Shot Scenario Evaluation}
We conduct zero-shot experiments with CLIP~\cite{radford2021learning} and word2vec~\cite{mikolov_efficient_2013} as language models.
PAPFZS3D~\cite{PAPFZS3D} uses prototype-based regression with text embeddings applied post-interaction, limiting their influence on predictions. 

In contrast, our zero-shot prototypes are initialized from the language model and refined through direct interaction, enabling better alignment between query tokens and text features. This preserves more semantic information for final similarity matching. We focus on validating zero-shot feasibility and do not compare with large-scale models like SegPoint~\cite{SegPoint} due to differences in training resources and data scale. Note that this experiment aims to validate the feasibility of zero-shot inference with our method. We do not compare against large-scale models such as SegPoint~\cite{SegPoint}, as there exists a significant disparity in training resources and the scale of utilized data. The results in Table~\ref{tb:zeroshot} demonstrate the efficacy and superiority of our proposed approach. 

\begin{table}[tbp]
    \centering
    \resizebox{\linewidth}{!}{\begin{tabular}{l l c c c c}
    \toprule
    Embed & Method & \makecell{2-way\\ 1-shot} & \makecell{2-way\\ 5-shot} & \makecell{3-way\\ 1-shot} & \makecell{3-way\\ 5-shot} \\
    \midrule 
    word2vec & 3DGenZ & 34.93 & 36.12 & 23.08 & 27.52 \\
    word2vec & PAPFZS3D & 59.98 & 63.54 & 48.91 & 55.62 \\
    CLIP & PAPFZS3D & 61.09 & 64.91 & 50.18 & 59.10 \\
    \rowcolor{lightblue!10}
    CLIP & \textbf{Ours} & \textbf{63.84} & \textbf{65.43} & \textbf{55.62} & \textbf{60.04} \\
    \bottomrule
    \end{tabular}}
    \caption{Zero-Shot evaluation result on the S3DIS dataset using mean-IoU criteria (\%). The best result of each column is highlighted with \textbf{bold font}.}
    \label{tb:zeroshot}
\end{table}
\begin{table}[t]
    \centering
    \resizebox{0.9\linewidth}{!}{\begin{tabular}{l c c c c c}
    \toprule
    ID &ProERA&LGPE &DRPE &Result & $\Delta$\\
    \midrule
    \uppercase\expandafter{\romannumeral1} & & &   &31.55 &\textcolor{black}{-41.53}\\
    \uppercase\expandafter{\romannumeral2} & &  & \checkmark  &64.84&\textcolor{black}{-8.24}\\
    \uppercase\expandafter{\romannumeral3} &   &\checkmark &  &60.22&\textcolor{black}{-12.86}\\
    \uppercase\expandafter{\romannumeral4} &\checkmark  &  &   &59.27&\textcolor{black}{-13.81}\\
    \uppercase\expandafter{\romannumeral5} &   &\checkmark &\checkmark   &70.48&\textcolor{black}{-2.60}\\
    \uppercase\expandafter{\romannumeral6} &\checkmark  &\checkmark  &  &69.35&\textcolor{black}{-3.73}\\
    \uppercase\expandafter{\romannumeral7} &\checkmark & &\checkmark  &70.17&\textcolor{black}{-2.91}\\
    \midrule
    \midrule
    ID& $\mathbf{p}_{text}$ &$\mathbf{p}_{dyn}$ &$\mathbf{p}_{raw}$ & Result &$\Delta$\\
    \midrule
     \uppercase\expandafter{\romannumeral8}& & & &   68.71 &\textcolor{black}{-4.37}\\
     \uppercase\expandafter{\romannumeral9}&\checkmark & & &   71.49 &\textcolor{black}{-1.59}\\
     \uppercase\expandafter{\romannumeral10}& &\checkmark & &   70.30 &\textcolor{black}{-2.78}\\
     \uppercase\expandafter{\romannumeral11}& & &\checkmark &   69.51 &\textcolor{black}{-3.57}\\
    \uppercase\expandafter{\romannumeral12}&\checkmark &\checkmark & &   71.75 &\textcolor{black}{-1.33}\\

    \uppercase\expandafter{\romannumeral13}&\checkmark & &\checkmark &   71.08 &\textcolor{black}{-2.00}\\
    \uppercase\expandafter{\romannumeral14}& &\checkmark &\checkmark &   70.56 &\textcolor{black}{-1.52}\\
    \midrule
    \rowcolor{lightblue!10}
    \uppercase\expandafter{\romannumeral15}&\checkmark &\checkmark &\checkmark &   \textbf{73.08} &-\\
    \bottomrule
    \end{tabular}}
    \caption{Ablation study of model components (\textbf{upper}) and prototypes (\textbf{lower}) on S3DIS $S^0$ with 2-way 1-shot.}
    \label{tb:ablation_c}
    \label{tb:ablation_p}
\end{table}



\subsection{Ablation Study}
Table~\ref{tb:ablation_c} highlights the effectiveness of each model component and each type of prototype, demonstrating our approach's ability to enhance prediction accuracy. The results show that dynamic prototypes have the most significant impact on performance. Table~\ref{tb:ablation_lossandproj} confirms that both $\mathcal{L}_{con}$ and $\mathcal{L}_{align}$ significantly improve performance. To evaluate the effectiveness of our proposed DRPE, we conduct an ablation study comparing it with learnable positional encoding (\textit{Learn PE} in the table)~\cite{wang2022simple} and sinusoidal~\cite{vaswani_attention_2017} positional encodings (\textit{Sin PE} in the table). As shown in Table~\ref{tb:ablation_lossandproj}, DRPE demonstrates clear advantages. Given that 3D point coordinates already provide accurate spatial positions, DRPE further enriches the representation by fusion query-support information.

\begin{figure}[t]
    \centering{
    \includegraphics[width=1\linewidth]{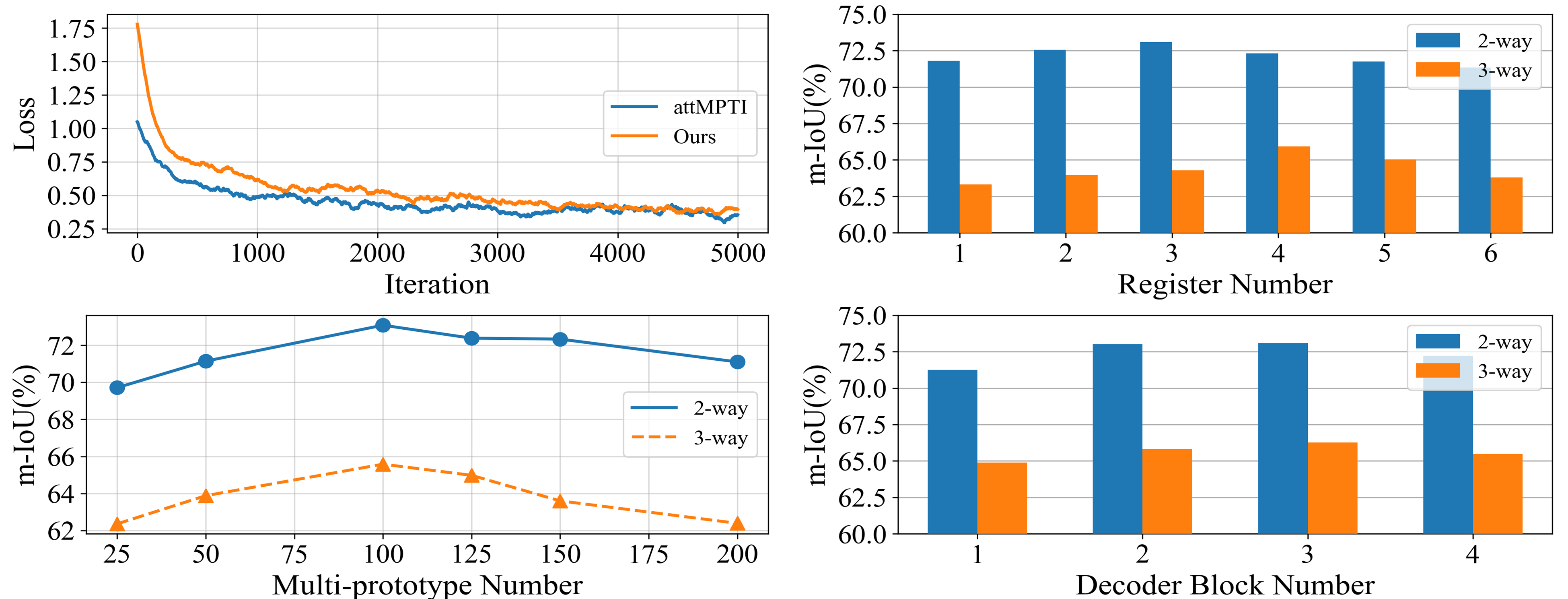}
    }
\caption{Visualized training loss curve (\textbf{upper left}). Ablation study results on number of registers  (\textbf{upper right}), decoder block (\textbf{lower right}), and multi-prototype (\textbf{lower left}).}
\label{fig:ablfig}
\end{figure}

Figure~\ref{fig:ablfig} presents loss curves of attMPTI and our model over the first 5,000 iterations, showing similar convergence speeds. Our model starts with a higher loss (first 3,000 iterations) due to its randomly initialized backbone, but achieves stable loss values comparable to a fully-supervised pre-trained method. Figure~\ref{fig:ablfig} also examines performance variations with different register and decoder block counts in 2-way 1-shot and 3-way 1-shot settings. As observed in~\cite{darcet2023vision}, register count requires careful tuning. We find $N+1$ registers work best for $N$-way $K$-shot tasks. Ablation on prototype and block numbers (Figure~\ref{fig:ablfig}) shows that 100 and 3 have the best performance-efficiency trade-off. 

 \begin{table}[tbp]
    \centering
    \begin{tabular}{l l c c}
    \toprule
    ID & Components &Result &$\Delta$\\
    \midrule
    \uppercase\expandafter{\romannumeral1} &$w/o$ $\mathcal{L}_{align}$ & 69.46 &\textcolor{black}{-4.24} \\
    \uppercase\expandafter{\romannumeral2} &$w/o$ $\mathcal{L}_{con}$ & 69.30 &\textcolor{black}{-4.50}\\
    \uppercase\expandafter{\romannumeral3} &$w/o$ $\mathcal{L}_{align}$+$\mathcal{L}_{con}$ & 68.55&\textcolor{black}{-5.25} \\
    \midrule
    \midrule
    \uppercase\expandafter{\romannumeral4} &Sin PE~\cite{vaswani_attention_2017} & 69.94 & \textcolor{black}{-3.14}\\
    \uppercase\expandafter{\romannumeral5} &Learn PE~\cite{wang2022simple} & 71.22 & \textcolor{black}{-1.86}\\
    \rowcolor{lightblue!10}
    \uppercase\expandafter{\romannumeral6} &\textbf{Ours} & \textbf{73.08 }&-\\
    
    \bottomrule
    \end{tabular}
    \caption{Ablation study of losses (upper) and positional encoding (lower) on S3DIS $S^0$.}
    \label{tb:app_ablpe}
    \label{tb:ablation_lossandproj}
\end{table}
\section{Conclusion}
We propose EPSegFZ, a pre-training-free 3D SemSeg model targeted for few-shot and zero-shot scenarios. A ProERA module is developed to enable the network to capture high-frequency features with less noise. Accurate query-prototype correspondence can be established using the proposed DRPE-based cross-attention. An LGPE module is designed to update prototypes with textual support information, fully exploit available data, reduce reliance on perfect visual information, and empower zero-shot inference. Furthermore, we designed a foreground-consistency alignment loss and a foreground-aware contrastive loss to effectively supervise the language-vision alignment and feature extraction.  

\section{Acknowledgement}
This research is supported by the National University of Singapore under the NUS College of Design and Engineering Industry-focused Ring-Fenced PhD Scholarship programme. It is also supported by the National Research Foundation (NRF) ``Centre for Advanced Robotics Technology Innovation (CARTIN)", and the National Robotics Programme (NRP) 2.0 funding initiative "Domain-specific Robotics Foundation Models for Manufacturing (DS-RFM). The authors would like to acknowledge useful discussions with Dr. Bruce Engelmann from Hexagon, Manufacturing Intelligence Division, Simufact Engineering GmbH.
\nolinenumbers
\bibliography{aaai2026}

\clearpage \appendix \twocolumn[
\begin{@twocolumnfalse}
\section*{\centering{Appendices for \\ {\emph{EPSegFZ: Efficient Point Cloud Semantic Segmentation for Few- and Zero-Shot Scenarios with Language Guidance}}}}
\end{@twocolumnfalse}
]
\section*{Overview}
This is the supplementary material, or can be called the appendix, of the paper ``EPSegFZ: Efficient Point Cloud Semantic Segmentation for Few- and Zero-Shot Scenarios with
Language Guidance.''  Firstly, we discuss the details of the evaluation metric we used for inference and the hyperparameter settings. Then, we illustrate the PyTorch-style pseudo-code of our EPSegFZ and provide more detailed explanations. Finally, we demonstrate the results of the extra experiment.

\section{Training Details and Hyperparameters}\label{app:evametric}
We train our model on two Nvidia RTX 3090 GPUs with 24GB of memory and test it on a single RTX 3090 GPU. Technically, our training process for one dataset can be done with one RTX 3090 GPU; the reason for using two GPUs is to accelerate the training on two datasets. All experiments are conducted with PyTorch 2.2.0 and CUDA 12.1 on an Ubuntu 22.04 operating system. The hyperparameter settings of our network and our AdamW optimizer are illustrated in Table~\ref{tb:hyp}. To preserve the generalization ability of our backbone and avoid overfitting, we assign a relatively small learning rate to the backbone compared to other modules. For fairness, the detailed setting in the backbone remains the same as previous works~\cite{zhao_few-shot_2021,PAPFZS3D}. The setting of hyperparameters for different network components is based on the best experimental result. We ran each test three times to obtain the final performance.
\begin{table}[ht]
    \centering
    \begin{tabular}{l c c}
    \toprule
     \makecell{Hyper-\\parameter} & Value  &\makecell{Tuning\\ range}\\
    \midrule
    Feature dimension   & 64 &32,64,96,128\\
    Backbone block number   & 3 &2,3,4\\
    Backbone $k$-NN number   & 20 &10,15,20,25\\
    Backbone learning rate &0.006 &0.005$\sim$0.01\\
    Backbone lr decay step &1000 &500$\sim$4000\\
    Backbone lr decay ratio &0.5 &0.4$\sim$0.8\\
    \midrule
    Decoder block number &3 &2,3,4\\
    Multi-prototype number &100 &\makecell{25,50,100,\\125,150,200}\\
Register number &3 &1,2,3,4,5,6\\
\midrule
    $\lambda_1^{*}$ &1.0 &0.1$\sim$1.0\\
      $\lambda_2^{*}$ &0.5&0.1$\sim$1.0\\
      $\lambda_3^{*}$ &0.7&0.1$\sim$1.0\\
     $\lambda_4^{*}$ &0.6&0.1$\sim$1.0\\
    \midrule
    $\tau$ & 0.5 &0.07$\sim$1.0\\
    $\lambda_{con}$ & 0.01 &0.01$\sim$0.5\\
    $\lambda_{align}$ & 0.02 &0.01$\sim$0.5\\
    \midrule
     Initial learning rate &0.001 &0.0001$\sim$0.005\\
     Decay step &5000 &2000$\sim$5000\\
     Decay ratio &0.5 &0.2$\sim$0.8\\
    \bottomrule
    \end{tabular}
    \caption{hyperparameter settings for training on S3DIS dataset with 2-way 1-shot task}
    \label{tb:hyp}
\end{table}

To align with the standard evaluation practices in the field of point cloud semantic segmentation, we assess the performance of all methods using the Mean Intersection-over-Union (m-IoU) metric, which is calculated as follows:
\begin{equation}
    \text{m-IoU} = \frac{1}{N}\sum_{c=1}^{N}\frac{TP^c}{TP^c+FP^c+FN^c},
\end{equation}
where the $TP^c$, $FP^c$, and $FN^c$ are the number of true positives, false positives, and false negatives, of the $c$-th class, respectively.

\begin{algorithm}[t]
    \caption{Pseudocode of the EPSegFZ decoder block forward function in PyTorch-style.}
    \definecolor{codeblue}{rgb}{0.25,0.5,0.5}
    \definecolor{codekw}{rgb}{0.85, 0.18, 0.50}
    \lstset{
      backgroundcolor=\color{white},
      basicstyle=\fontsize{7.5pt}{7.5pt}\ttfamily\selectfont,
      columns=fullflexible,
      breaklines=true,
      captionpos=b,
      commentstyle=\fontsize{7.5pt}{7.5pt}\color{codeblue},
      keywordstyle=\fontsize{7.5pt}{7.5pt}\color{codekw},
    }
    \begin{lstlisting}[language=python]
    # qx: query feature (N, M, C)
    # sx: support feature (N, K, M, C)
    # sy: support mask (N, K, M)
    # st: text prompt of support classes with the shape (N+1,*)
    def forward(qx,sx,sy,st):
        # Multi-prototype sampling
        # Only adopted in the first decoder block
        mul_proto = MPS(sx,sy) # (N+1,np,C)
        # Get the prototypes with average pooling
        raw_proto = torch.mean(mul_proto,1) #(N+1,C)
        # Initialized prototype tokens
        p_token = raw_proto.clone()
        # Process with the ProERA module
        qx = ProERA(qx,q_reg,p_token) 
        # q_reg and s_reg are registers
        mul_proto = ProERA(mul_proto,s_reg,p_token)
        # Update prototype tokens in LGPE
        text_proto = CLIP(st) #(N+1,C)
        #dynamic prototypes
        dyn_proto = mask_avg_pool(mul_proto) 
        
        #W=[lambda_1 to lambda_4] (1,4)
        p_token = torch.mm(W,torch.cat(p_token,\
            raw_proto,dyn_proto,text_proto))
            
        # Calculating the DRPE value
        dE = pairwise distance(qx.view(-1,C),pTok)
        dC = torch.mm(qx/qx.norm(),p_token/p_token.norm())
        R = SinEmb(dE)+SinEmb(dC) # (M,N+1,C)
        qx = cra(qx,p_token,p_token,R)
        p_token = cra(p_token,qx,qx,R.T)
        return qx,p_token
    \end{lstlisting}
    \label{algo:pseudocode}
\end{algorithm}
\begin{figure}[!t]
    \centering
    \includegraphics[width=0.9\linewidth]{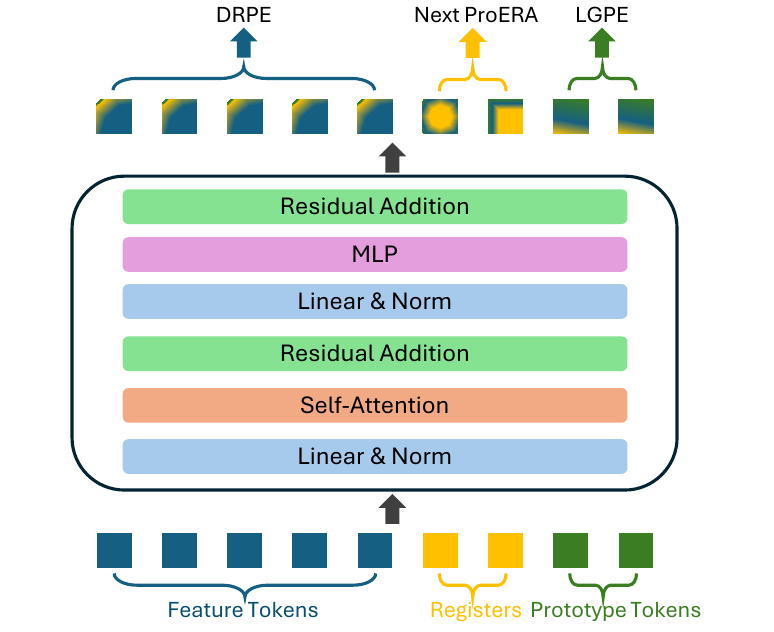}
    \caption{Visualized structure of our ProERA module, the output registers will be directly preserved for the ProERA module in the next decoder block.}
    \label{fig:ProERAstructure}
\end{figure}
\section{More Explanations}\label{app:pesudocode}
\subsection{Selected Baselines}
We select several baselines for quantitative comparison. Fine-tuning~\cite{shaban2017one} is selected as the one without a few-shot setting. This model represents that we use some of (equals to the shot number) the test sample to fine-tune a pre-trained model, and then conduct the prediction. It is not surprising that such a method suffers from low generalization ability. The success cases of it mainly come from a well-trained encoder that can identify base classes. ProtoNet~\cite{prototypenet} is a representative method of classic few-shot learning. It first pre-trains the encoder, then generates one prototype for each support class and makes a prediction. attMPTI~\cite{zhao_few-shot_2021} made a breakthrough, which generates multiple representative prototypes through learning geometric and semantic features with self-attention. After that, it uses a label propagation~\cite{iscen_label_2019} method to conduct the prediction. CWT~\cite{lu2021simpler} and 2CBR~\cite{2CBR} introduce additional classifiers to boost the classification of novel classes. PEFC~\cite{zhang2023PEFC} and QGE~\cite{QGE} dig deep into the query-support relationship for better similarity measuring. PAPFZS3D~\cite{PAPFZS3D} utilizes text features to improve support representations. However, it does not allow the features to fuse together and only uses the text feature as auxiliary supervision. Seg-NN/Seg-PN~\cite{segpn} proposed a method to avoid pre-training. Nonetheless, as mentioned in the Introduction section, it lacks sufficient care for high-frequency information. SDSimPoint~\cite{wang2025sdsimpoint} decoupled the similarity measuring into shallow and deep parts. Although it achieves promising results and interoperability, the heavy structure and dependency on pre-training limit its application.

\subsection{Visualized Architecture and Pseudocode}
We provide a PyTorch-style pseudocode of the forward function in our EPSegFZ decoder block, as shown in Algorithm~\ref{algo:pseudocode}.

\begin{figure}[t]
    \centering
    \includegraphics[scale=0.37]{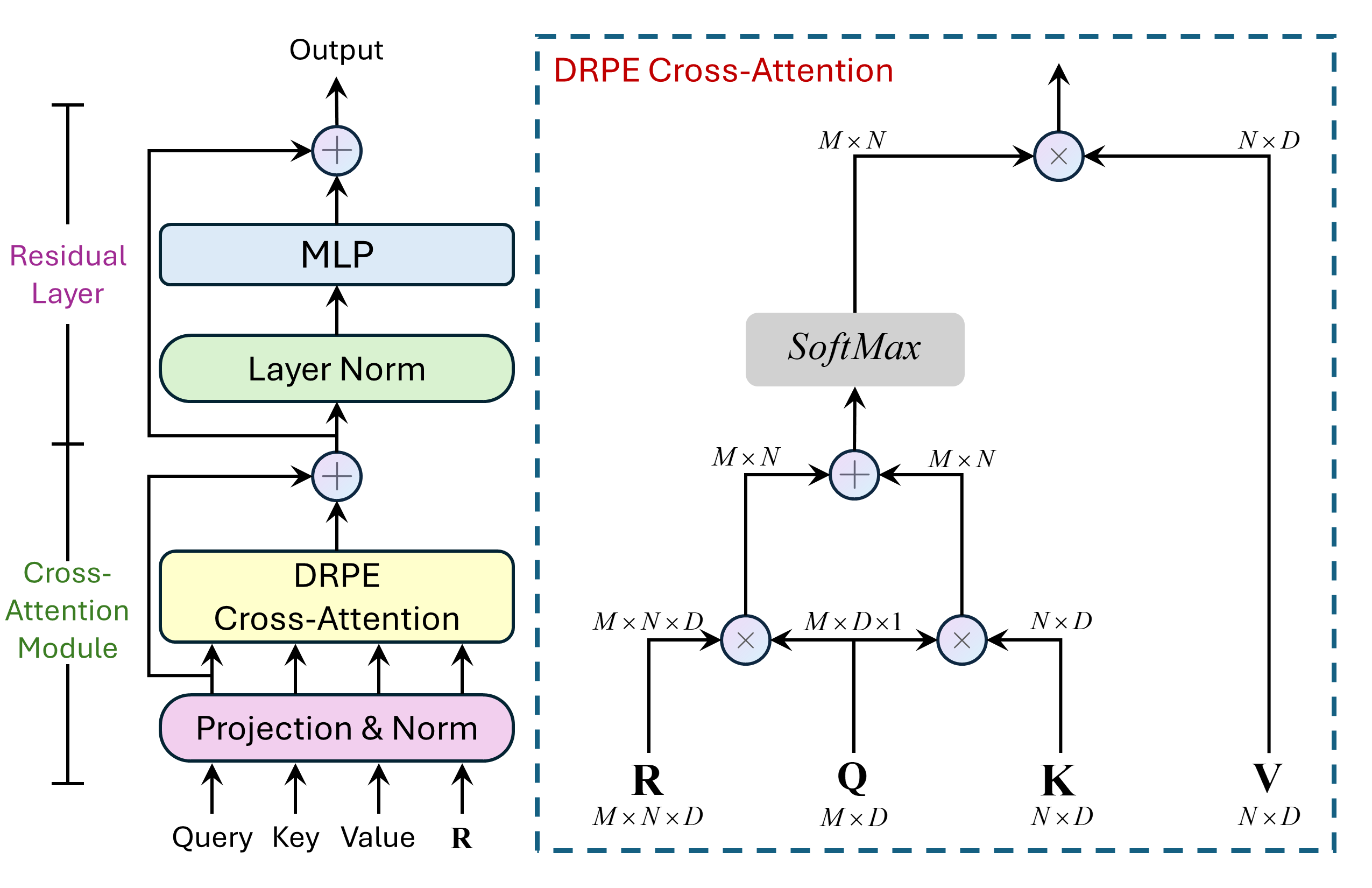}
    \caption{The architecture of our decoder layer. The visualized calculation process of the DRPE-based cross-attention is illustrated on the right.}
    \label{fig:rpecrossattention}
\end{figure}

The visualized architecture of our ProERA module is illustrated in Figure~\ref{fig:ProERAstructure}. Figure~\ref{fig:rpecrossattention} depicts the structure of our DRPE Cross-attention. The right half of the figure shows how we integrate the positional embedding into query tokens. For the $i$-th block, the outputs can be written as:
\begin{equation}
    \begin{aligned}
    \hat{\mathbf{X}}_q^i &= \text{Res}\left( \text{CRA}(\Tilde{\mathbf{X}_q^i},\mathbf{p}^i,\mathbf{p}^i,\mathbf{R}^i)\right) ,\\
        \hat{\mathbf{p}}^i &=\text{Res}\left( \text{CRA}(\mathbf{p}^i,\Tilde{\mathbf{X}_q^i},\Tilde{\mathbf{X}_q^i},(\mathbf{R}^i)^{\top})\right),
    \end{aligned}
\end{equation}
where  $\text{CRA}(Q,K,V,R)$ is our DRPE cross-attention module.
\subsection{Sinusoidal Embedding}
The $\textit{SinEmb}(\cdot)$ function is defined as:
\begin{equation}
    \begin{aligned}
    \textit{SinEmb}(l)=[&sin(\frac{l}{\gamma^{\frac{0}{D}}}),cos(\frac{l}{\gamma^{\frac{0}{D}}}),\cdots \\&\cdots,sin(\frac{l}{\gamma^{\frac{D-2}{D}}}),cos(\frac{l}{\gamma^{\frac{D-2}{D}}})  ],
    \end{aligned}
\end{equation}
where $l$ is a postional index, $\gamma$ is a scale parameter which is set to 10000~\cite{vaswani2017attention}. 
\subsection{Detail of Formulations}
As discussed in the Methodology section, the foreground semantics are supervised in a contrastive manner. Specifically, we calculate the point-wise similarity only among foreground objects, as foreground instances may appear as backgrounds in different scenes. Pulling all background pairs closer indiscriminately could introduce significant noise. Mathematically, the proposed loss can be expressed as:
\begin{equation}
    \begin{aligned}
    \mathcal{L}_{con}=\sum_{(i, j)} & -\frac{1}{2} \log \frac{\exp \left(\mathbf{x}_s^{i} \cdot \mathbf{x}_q^{j} / \tau\right)}{\sum_{n=1}^N \exp \left(\mathbf{x}_s^{i} \cdot \mathbf{x}_q^{n} / \tau\right)} \\
    & -\frac{1}{2} \log \frac{\exp \left(\mathbf{x}_s^{i} \cdot \mathbf{x}_q^{j} / \tau\right)}{\sum_{n=1}^N \exp \left(\mathbf{x}_s^{n} \cdot \mathbf{x}_q^{j} / \tau\right)},
    \end{aligned}
    \label{eq:lcon}
\end{equation}
where $\mathbf{x}_q^{j}$ is the $j$-th point in query point cloud and $N$ is the total number of points.



\section{More Experimental Results}\label{app:extraexp}
\begin{table}[t]
    \tabcolsep=0.3cm
    \centering
    \resizebox{1.0\linewidth}{!}{\begin{tabular}{l c c c c c}
    \toprule
    ID &Registers& Prototype tokens& $\mathbf{R}_E$& $\mathbf{R}_C$& Result\\
    \midrule
    
    \uppercase\expandafter{\romannumeral1} & & & & &  65.91\\
    \uppercase\expandafter{\romannumeral2} & &\checkmark & & & 67.14\\
    \uppercase\expandafter{\romannumeral3} &\checkmark& & & &  66.49\\
    \uppercase\expandafter{\romannumeral4} &\checkmark&\checkmark & & &  67.53\\
    \cmidrule{1-6}
    
    \uppercase\expandafter{\romannumeral5} & & &\checkmark & &  66.40\\
    \uppercase\expandafter{\romannumeral6} & & & &\checkmark &  66.67\\
    \uppercase\expandafter{\romannumeral7} & & &\checkmark &\checkmark & 67.25\\
    \cmidrule{1-6}
    \uppercase\expandafter{\romannumeral8} & &\checkmark &\checkmark & & 68.06\\
    \uppercase\expandafter{\romannumeral9} & &\checkmark & &\checkmark & 68.13\\
    \uppercase\expandafter{\romannumeral10} & &\checkmark &\checkmark &\checkmark & 68.95\\
    \cmidrule{1-6}
    \uppercase\expandafter{\romannumeral11} & \checkmark& &\checkmark & & 68.88\\
    \uppercase\expandafter{\romannumeral12} &\checkmark& & &\checkmark & 68.69\\
    \uppercase\expandafter{\romannumeral13} &\checkmark& &\checkmark &\checkmark & 69.24\\
    \cmidrule{1-6}
    \uppercase\expandafter{\romannumeral14} & \checkmark&\checkmark &\checkmark & & 72.92\\
    \uppercase\expandafter{\romannumeral15} &\checkmark&\checkmark & &\checkmark & 72.36\\
    \rowcolor{lightblue!10}\uppercase\expandafter{\romannumeral16} &\checkmark&\checkmark &\checkmark &\checkmark & \textbf{73.08}\\
    \toprule
    \end{tabular}}
    \caption{Ablation study of detailed components on S3DIS $S^0$ with 2-way 1-shot}
    \label{tb:ablation_detail}
\end{table}

\begin{table*}[t]
    \centering
   \resizebox{\linewidth}{!}{ \begin{tabular}{l c c c c c c c c c c c c}
    \toprule
    \multirow{4}{*}{ Method } & \multicolumn{6}{c}{ 1-way } & \multicolumn{6}{c}{ 2-way } \\
    \cmidrule(lr){2-7} \cmidrule(lr){8-13} 
    & \multicolumn{3}{c}{ 1-shot } & \multicolumn{3}{c}{ 5-shot } & \multicolumn{3}{c}{ 1-shot } & \multicolumn{3}{c}{ 5-shot } \\
    \cmidrule(lr){2-4}  \cmidrule(lr){5-7} \cmidrule(lr){8-10} \cmidrule(lr){11-13} 
    & $S^0$ & $S^1$ & mean & $S^0$ & $S^1$ &  \multicolumn{1}{c}{mean} & \multicolumn{1}{c}{$S^0$} & $S^1$ & mean & $S^0$ & $S^1$ & mean \\
    \midrule
    AttMPTI~\cite{zhao_few-shot_2021} &36.32 &38.36 &37.34 &46.71 &42.70 &44.71 &31.09 &29.62 &30.36 &39.53 &32.62 &36.08 \\
    QGE~\cite{zhang2023PEFC} &41.69 &39.09 &40.39 &50.59 &46.41 &48.50 &33.45 &30.95 &32.20 &\underline{40.53} &36.13 &{38.33}\\
    PAPFZS3D~\cite{PAPFZS3D} &35.50 &35.83 &35.67 &38.07 &39.70 &38.89 &25.52 &26.26 &25.89 &30.22 &32.41 &31.32\\
    COSeg~\cite{an2024rethinking} &\underline{46.31} &\textbf{48.10} &\textbf{47.21} &\textbf{51.40} &\underline{48.68} &\underline{50.04} &\textbf{37.44} &\textbf{36.45} &\textbf{36.95} &\textbf{42.27} &\textbf{38.45} &\textbf{40.36}\\
   \rowcolor{lightblue!10} \textbf{Ours} &\textbf{47.45} &\underline{46.86} &\underline{47.16} &\underline{50.77} &\textbf{49.52} &\textbf{50.15} &\underline{34.35} &\underline{32.18} &\underline{33.27} &{39.82} &\underline{37.37} &\underline{38.60}\\
    \toprule
    \end{tabular}}
    \caption{Evaluation result on the S3DIS dataset using mean-IoU criteria (\%) under the COSeg's setting. The best result of each column is highlighted with \textbf{bold font}, and the second best is noted with \underline{underline}.}
    \label{tb:coseg}
\end{table*}
\begin{table}[!t]
    \centering
    \begin{tabular}{c c c c}
    \toprule
    Model & $\#$Params.& Training Time& GPUs\\
    \midrule
        AttMPTI &352.19K &12$\sim$16h &1\\
        COSeg &7.68M &27$\sim$31h &4\\
        Ours &2.02M	&4$\sim$6h	&1\\
    \bottomrule
    \end{tabular}
    \caption{Model Complexity and Training Usage for the 2-way 1-shot task.}
    \label{tb:parameters}
\end{table}
\subsection{Ablation of Individual Components} Table~\ref{tb:ablation_detail} shows the effectiveness of each component in our model. The prototype tokens and registers column means using registers or prototype tokens in our ProERA module. Although the prototype tokens are not utilized in the ProERA module in some individual experiments, the LGPE module will still update the prototypes without the learned tokens. For model performance without the LGPE module, Exp.\uppercase\expandafter{\romannumeral1} in Table~\ref{tb:ablation_p} denotes the corresponding result. The results in the tables indicate that the absence of prototype tokens significantly diminishes the performance, thereby underscoring the crucial role of the interaction between prototypes and query features.

\begin{figure}[ht]
    \centering
    \includegraphics[width=1\linewidth]{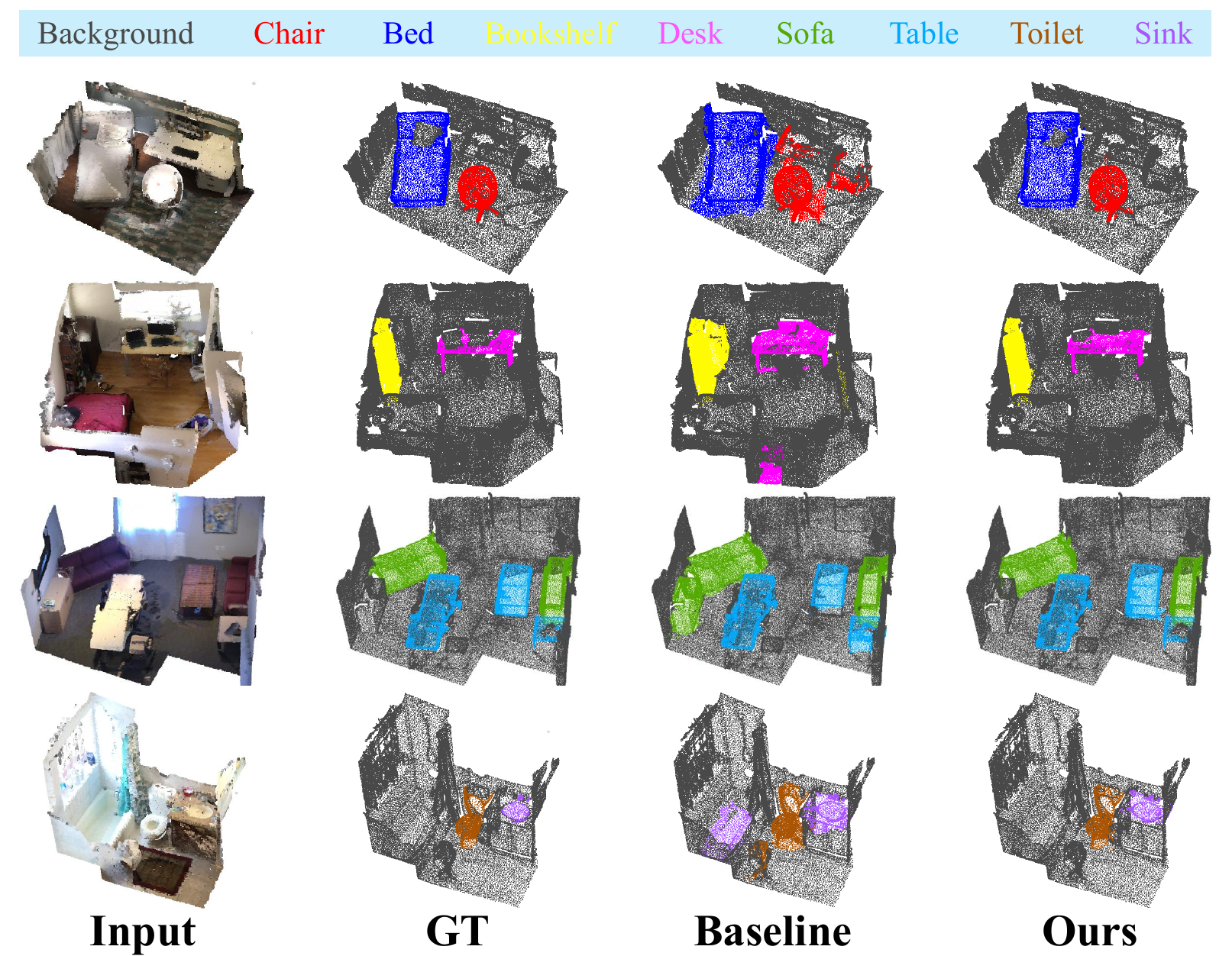}
    \caption{Visualized segmentation result and comparison on the ScanNet dataset}
    \label{fig:scannetres}
\end{figure}
\begin{figure}[ht]
    \centering
    \includegraphics[width=1.0\linewidth]{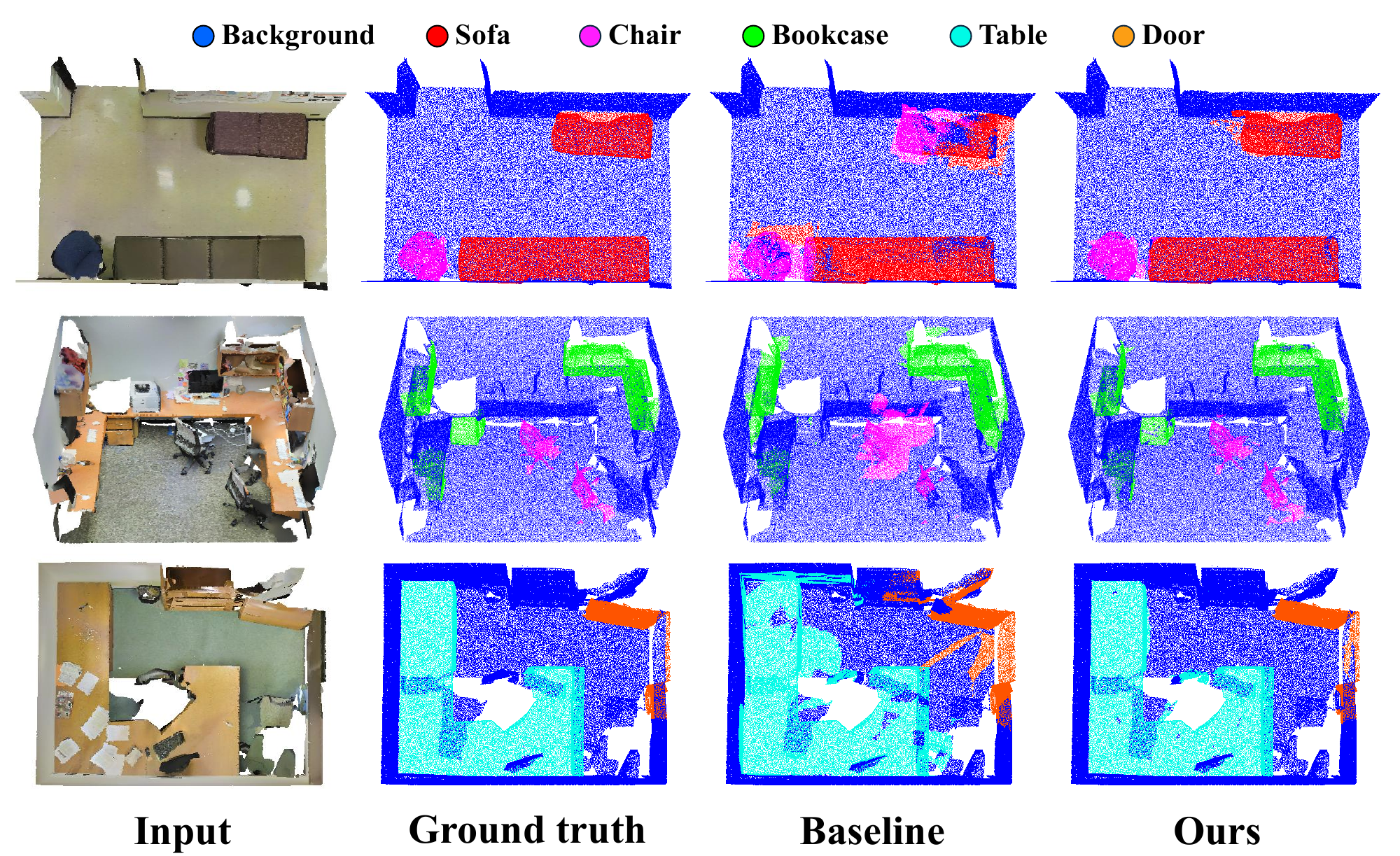}
    \caption{Visualized segmentation result and comparison on the S3DIS dataset}
    \label{fig:ablviz}
\end{figure}
\begin{figure}[ht]
    \centering
    \includegraphics[width=1.0\linewidth]{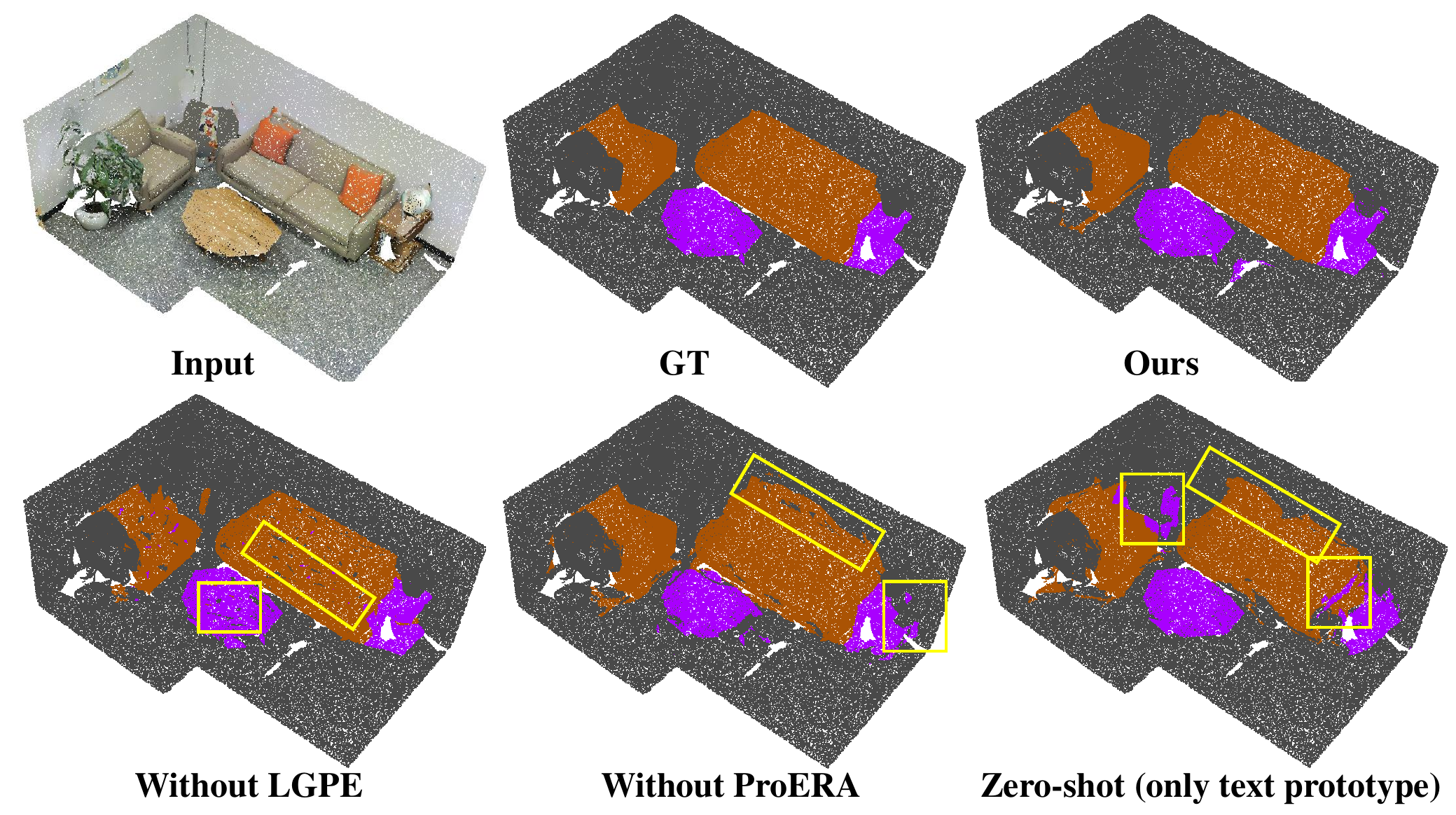}
    \caption{Visualiation of results on S3DIS with various settings.}
    \label{fig:extraviz}
\end{figure}
\begin{figure}[ht]
    \centering
    \includegraphics[width=0.9\linewidth]{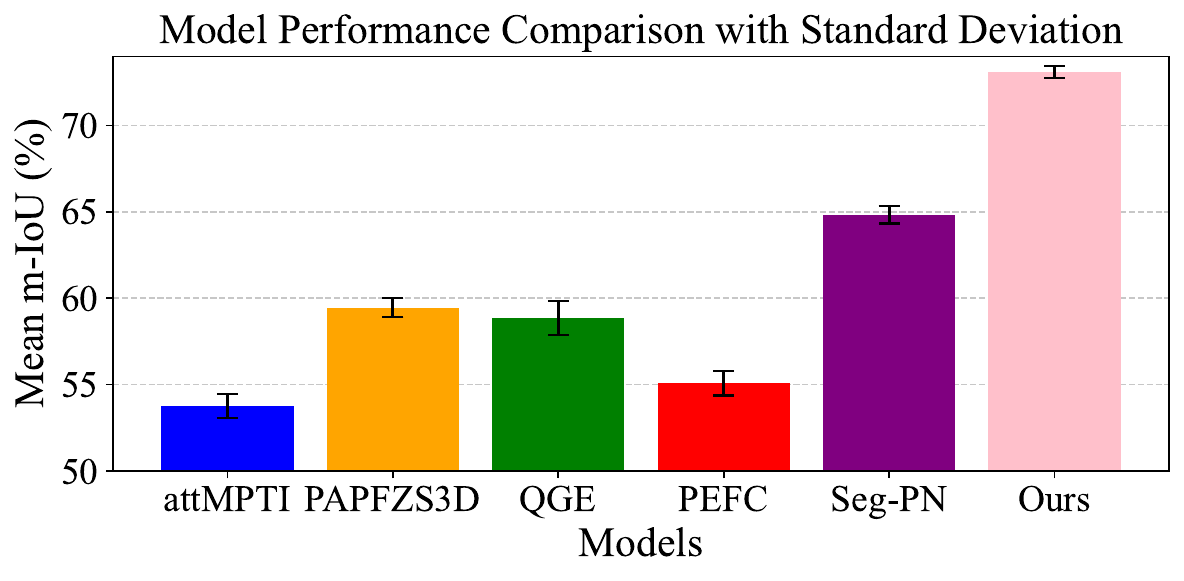}
    \caption{Comparison of mean m-IoU performance across various models on S3DIS. Error bars represent the standard deviation across three individual tests.}
    \label{fig:std}
\end{figure}
\subsection{More Visualized Result}\label{sec:app_viz} 
As claimed in the previous Experiment section, we visualized the segmentation result on the ScanNet~\cite{scannet} dataset. The figures show that compared to Baseline~\cite{segpn}, our method performs better in precisely segmenting target classes. We additionally investigate the performance of our method and Seg-PN~\cite{segpn} in S3DIS~\cite{s3dis}. Figure~\ref{fig:ablviz} illustrates the segmentation result. The test environment is a whole block. In this case, Seg-PN~\cite{segpn}, as it mainly focuses on low-frequency information and lacks the external instruction of LLM, performs worse than our method. Furthermore, to perceive the influence of each module directly, we visualized the results in different settings in Figure~\ref{fig:extraviz}. Without the proposed ProERA, the misclassification rate of points near object boundaries increases significantly. The absence of LGPE further results in poor spatial consistency within object regions. Moreover, in the zero-shot setting, performance degrades compared to the few-shot setting due to the lack of visual support samples. Figure~\ref{fig:std} presents a comparative analysis of the mean m-IoU (mean Intersection over Union) performance across six different models, including our proposed model. The x-axis represents the models being compared, while the y-axis shows the mean m-IoU values in percentage (\%). Error bars are included to indicate the standard deviation for each model's performance. The results clearly demonstrate that our proposed model achieves superior performance compared to existing methods, with a high m-IoU and relatively low variability as indicated by the small error bars. This suggests that our model not only performs well on average but also exhibits robustness across different runs or datasets.

\subsection{Different Benchmark Settings.}~\label{sec:supexpnewbenchmark}
COSeg~\cite{an2024rethinking} proposes a new benchmark setting for 3D few-shot semantic segmentation. Specifically, it uses more sparse sampling with an unfixed number of points to conduct the few-shot episodic learning paradigm. For convenience, we arbitrarily sample 20480 points sparsely using the method in COSeg. During our training, we use the DGCNN as the backbone. Table~\ref{tb:coseg} illustrates the results under the new benchmark setting. We can see that in the 1-way few-shot task, our model shares similar performance with the SOTA designed specifically for this benchmark and outperforms the previous SOTA used in general settings. However, for 2-way few-shot tasks, our method is slightly lower than the COSeg. There are two possible reasons: 1) the COSeg adopts a pre-trained Stratified Transformer~\cite{lai_stratified_2022}, which is far more powerful than our DGCNN that trains together with the subsequent networks, and 2) COSeg utilizes memory-bank-like base prototypes to store the knowledge with a cascade network, which preserves more information but requires a lot of memory. As claimed in COSeg, they require 4 RTX 3090 GPUs for training and testing, while our EPFSeg only requires 1 GPU for both training and testing. Thus, in resource-intensive scenarios, our method can be deployed more efficiently than COSeg. Table~\ref{tb:parameters} illustrates the number of trainable parameters and resources used in training. The COSeg has more parameters than ours and suffers from a long training time. The attMPTI~\cite{zhao_few-shot_2021}, though it contains fewer trainable parameters, spends notable time on graph construction. Our methods not only save pre-training time but also reduce the usage of GPUs, which shows the advantage of efficiency. 
\begin{table}[t]
    \centering
    
    \resizebox{\linewidth}{!}{\begin{tabular}{c c c c c c}
    \toprule
    ID& Method &Backbone&Pre-training& Result\\
    \midrule
    \uppercase\expandafter{\romannumeral1}&  \textbf{Ours} & KPConv &\XSolidBrush & 65.57\\
    \uppercase\expandafter{\romannumeral2}& \textbf{Ours} & PointNet++ &\XSolidBrush & 63.92 \\
    \uppercase\expandafter{\romannumeral3}& \textbf{Ours} & Point-NN &\XSolidBrush & 72.31 \\
    \midrule
    \uppercase\expandafter{\romannumeral4}&ProtoNet & DGCNN &\XSolidBrush & 30.54(\textcolor[RGB]{255,0,0}{$\downarrow$17.85})\\
    \uppercase\expandafter{\romannumeral5}&attMPTI & DGCNN &\XSolidBrush & 38.69(\textcolor[RGB]{255,0,0}{$\downarrow$15.08})\\
    \uppercase\expandafter{\romannumeral6}&PEFC & DGCNN &\XSolidBrush & 47.02(\textcolor[RGB]{255,0,0}{$\downarrow$8.07})\\
    \uppercase\expandafter{\romannumeral7}&PAPFZS3D & DGCNN &\XSolidBrush & 47.83(\textcolor[RGB]{255,0,0}{$\downarrow$11.62})\\
    \rowcolor{lightblue!10}\uppercase\expandafter{\romannumeral8}&\textbf{Ours} & DGCNN &\XSolidBrush &\textbf{73.08} \\
    \midrule
    \uppercase\expandafter{\romannumeral9}&ProtoNet& DGCNN &\checkmark & 48.39\\
    \uppercase\expandafter{\romannumeral10}&attMPTI & PointNet++ &\checkmark &47.13\\
    \uppercase\expandafter{\romannumeral11}&attMPTI & DGCNN &\checkmark &53.77\\
    \uppercase\expandafter{\romannumeral12}&PEFC & DGCNN &\checkmark &55.09\\
    \uppercase\expandafter{\romannumeral13}&PAPFZS3D & DGCNN &\checkmark &59.45\\
    \uppercase\expandafter{\romannumeral14}&\textbf{Ours} & DGCNN &\checkmark& 73.40(\textcolor[RGB]{0,200,0}{$\uparrow$0.02})\\
    \bottomrule
    \end{tabular}}
    \caption{Ablation Study of Backbones and Pre-trainings On S3DIS $S^0$ With 2-way 1-shot}
    \label{tb:traintype}
\end{table}
\subsection{Backbones and Pre-Training.}\label{sec:app_backbone} To investigate the influence of backbones, we conduct experiments that substitute DGCNN~\cite{wang_dynamic_2019} with KPConv~\cite{thomas_kpconv_2019} and PointNet++~\cite{qi_pointnet_2017}, respectively. The corresponding results are shown in Table~\ref{tb:traintype}. We can see that although we use PointNet++, a backbone that is considered weaker than DGCNN, our method can still obtain satisfactory performance (surpasses the attMPTI). Moreover, the experiments with pre-training in Table~\ref{tb:traintype} follow the paradigm and settings in~\cite{zhao_few-shot_2021} to pre-train the backbone. The results illustrate that methods like attMPTI encounter severe performance decline without pre-training. It also shows that with a pre-training like~\cite{zhao_few-shot_2021}, our model has limited improvement because of the class bias imported by the pre-training dataset. For attMPTI, the advantages of such pre-training outweigh its drawbacks, as it provides ample feature extraction capabilities for the backbone. Our method yields robust results, regardless of the existence of pre-training.


\begin{figure}[!ht]
    \centering
    \subfloat{}{
        \includegraphics[width=1\linewidth]{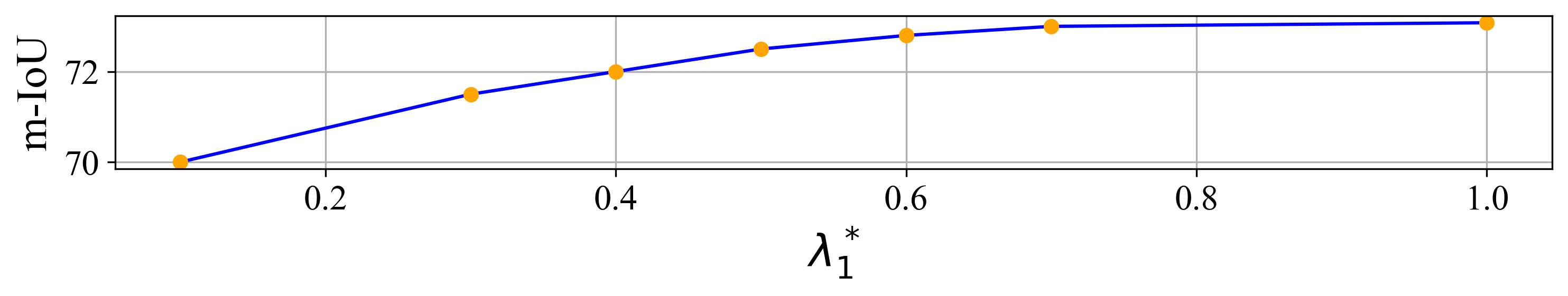}
    }
    \subfloat{}{
        \includegraphics[width=1\linewidth]{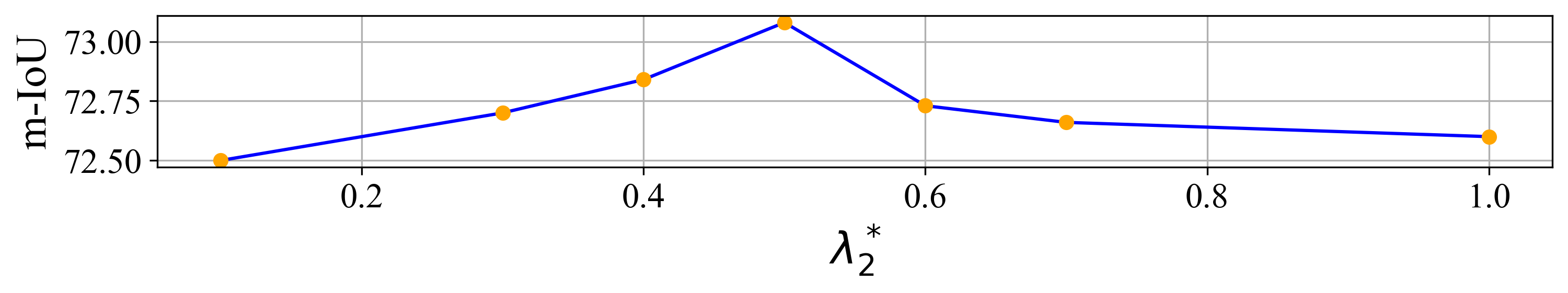}
    }
    \subfloat{}{
        \includegraphics[width=1\linewidth]{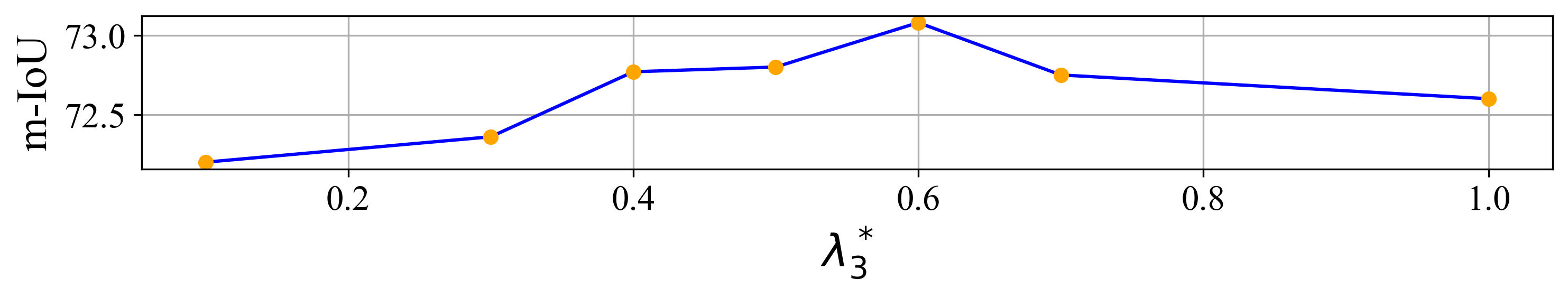}
    }
    \subfloat{}{
        \includegraphics[width=1\linewidth]{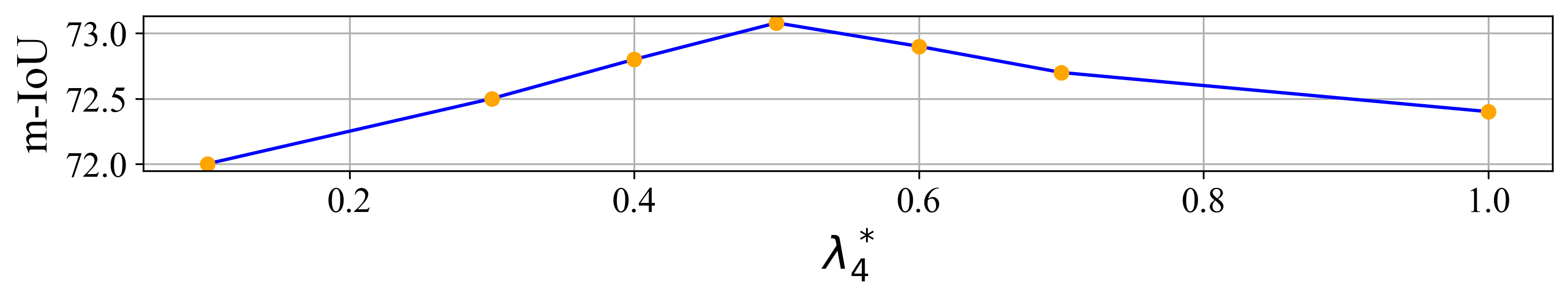}
    }
    \caption{Sensitive analysis of $\lambda_i^{*}; i\in[1,4]$. The result shows that although the performance does not vary too much, our chosen value can be considered as an optimal selection.}
    \label{fig:sensitive_analy}
\end{figure}

\subsection{Sensitive Analysis}
\begin{figure}[!t]
\centering
    \subfloat[]{
        \includegraphics[width=4cm]{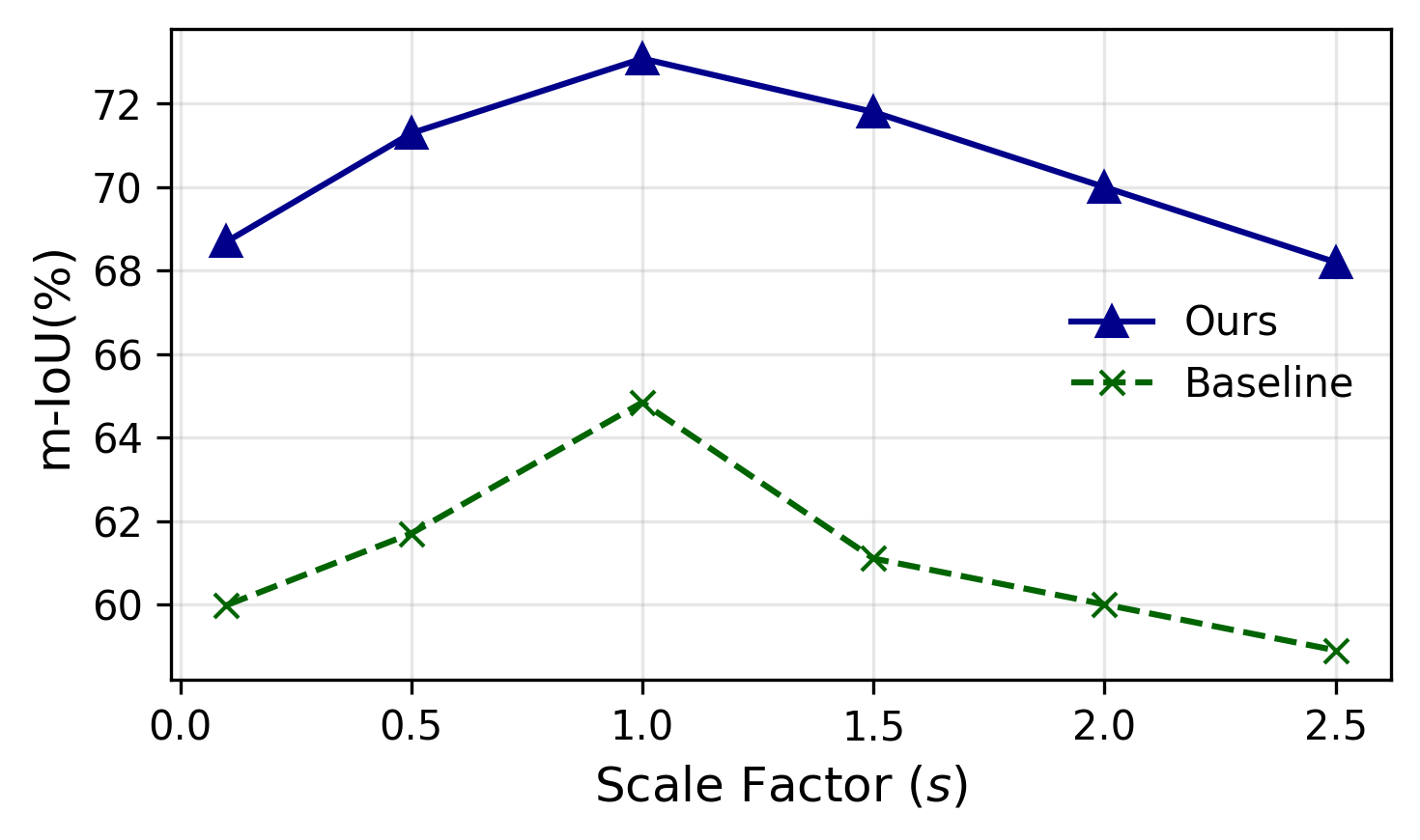}
    }
    \subfloat[]{
        \includegraphics[width=4cm]{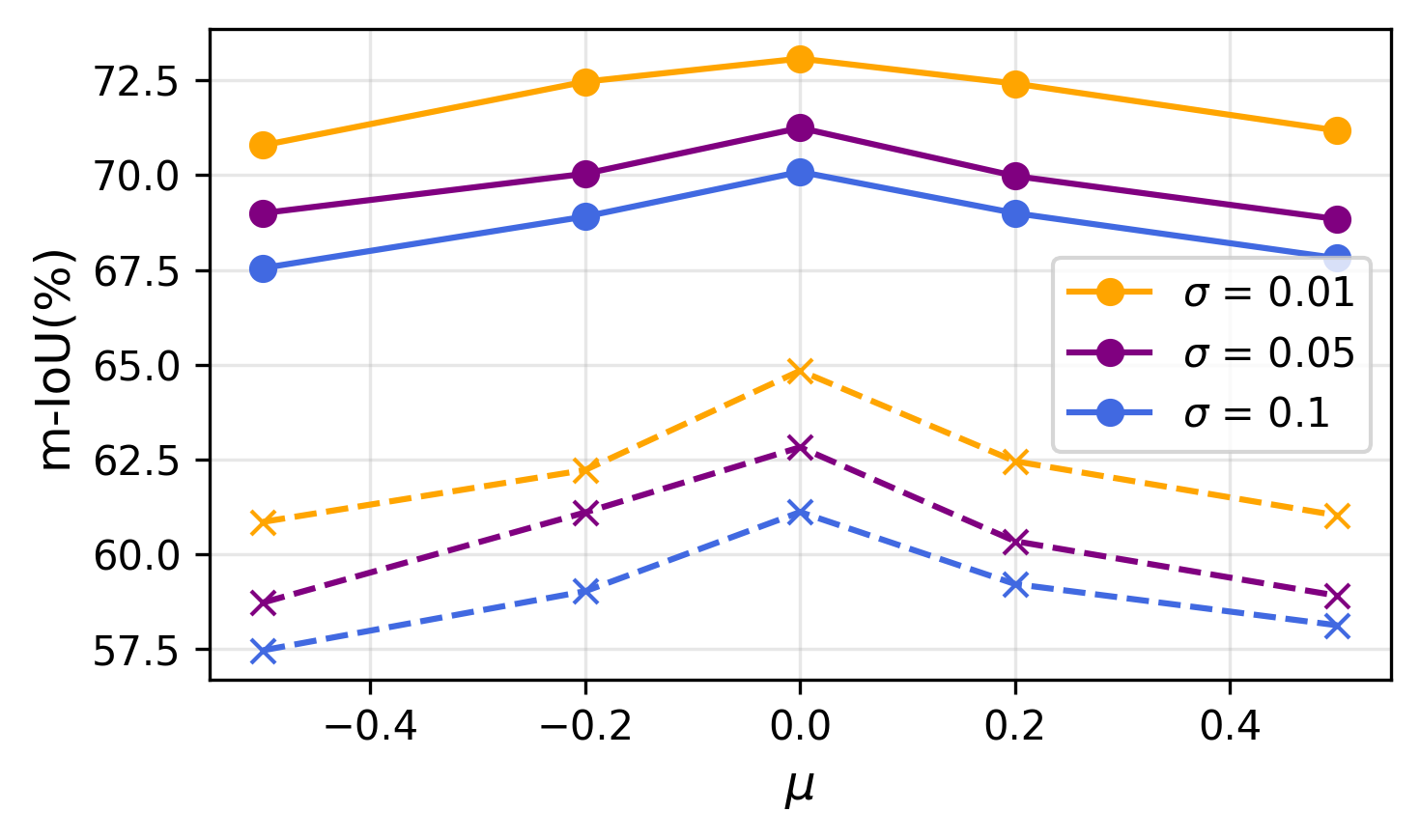}
    }
\caption{Performance under different (a) scaling factor $s$ and (b) Gaussian point-wise noise. The solid line represents our method, while the dashed line refers to the baseline.}
\label{fig:noise}
\end{figure}

In this subsection, we analyze the sensitivity of the $\lambda_i^{*};i\in[1,4]$ and the noise. Figure~\ref{fig:sensitive_analy} illustrates the result of each test from top to bottom. We can find that $\lambda_1^{*}$ has a more serious influence on the performance. This proves the importance of well-learned prototype tokens; if such information is not sufficient for the final prediction, the performance can decrease. The result of $\lambda_4^{*}$ shows that too little or too much language guidance can both result in performance decline. Therefore, an effective balancing strategy, such as a learnable MLP with dynamic weighting, is desirable. In this work, we adopt a projection MLP to align the text and vision representations while simultaneously achieving a mutual balance between the two modalities. For noise, we added a Gaussian-based point-wise jitter and a scale noise $s$ to each test sample, i.e. $\mathbf{P}_{test} = s \cdot (\mathbf{P}_{raw} + z)$ where $z\sim  \mathcal{N}(\mu,\sigma^2)$. Figure~\ref{fig:noise} illustrates the performance under different settings of $s$ and $z$.  The figure demonstrates that our method is more robust than the baseline, showing a slower decline in performance as the noise intensity increases.

\begin{table}[!t]
    \centering
    \resizebox{\linewidth}{!}{\begin{tabular}{c c c c c c c c}
        \toprule
        \multirow{2}{*}{Method} & \multicolumn{3}{c}{2-Way 1-Shot} & \multicolumn{3}{c}{2-Way 5-Shot}\\
        \cmidrule(l){2-4} \cmidrule(l){5-7}
        &$S^0$& $S^1$ & Mean  &$S^0$ & $S^1$ & Mean\\
        \midrule
        Fine-Tuning &36.48& 39.54 & 36.24 &41.79 &43.63 &42.71   \\ 
        ProtoNet  &39.64 & 41.19& 40.42 &43.08&44.12&43.60  \\
        attProtoNet  &42.95 & 43.70& 43.33 &45.01&46.97&45.99  \\
        MPTI    &40.27& 43.15& 44.21 &45.37&46.11&45.74  \\
        attMPTI &44.93& 49.23&47.08&51.32 &53.02&52.17  \\
        Wang et.al.&47.76&50.83&49.30&52.07&54.11&53.09  \\
        \rowcolor{lightblue!10}\bf{Ours} &\bf{52.73} &\bf{54.51} &\bf{53.62} &\bf{57.09}&\bf{56.44}&\bf{56.77}\\  
    \bottomrule
    \end{tabular}}
    \caption{{Quantative Results on the EIF 3D Dataset (\%)}}
    \label{tb:phcfres}
\end{table}

\begin{figure}[t]
  \centering
  \includegraphics[width=1.0\linewidth]{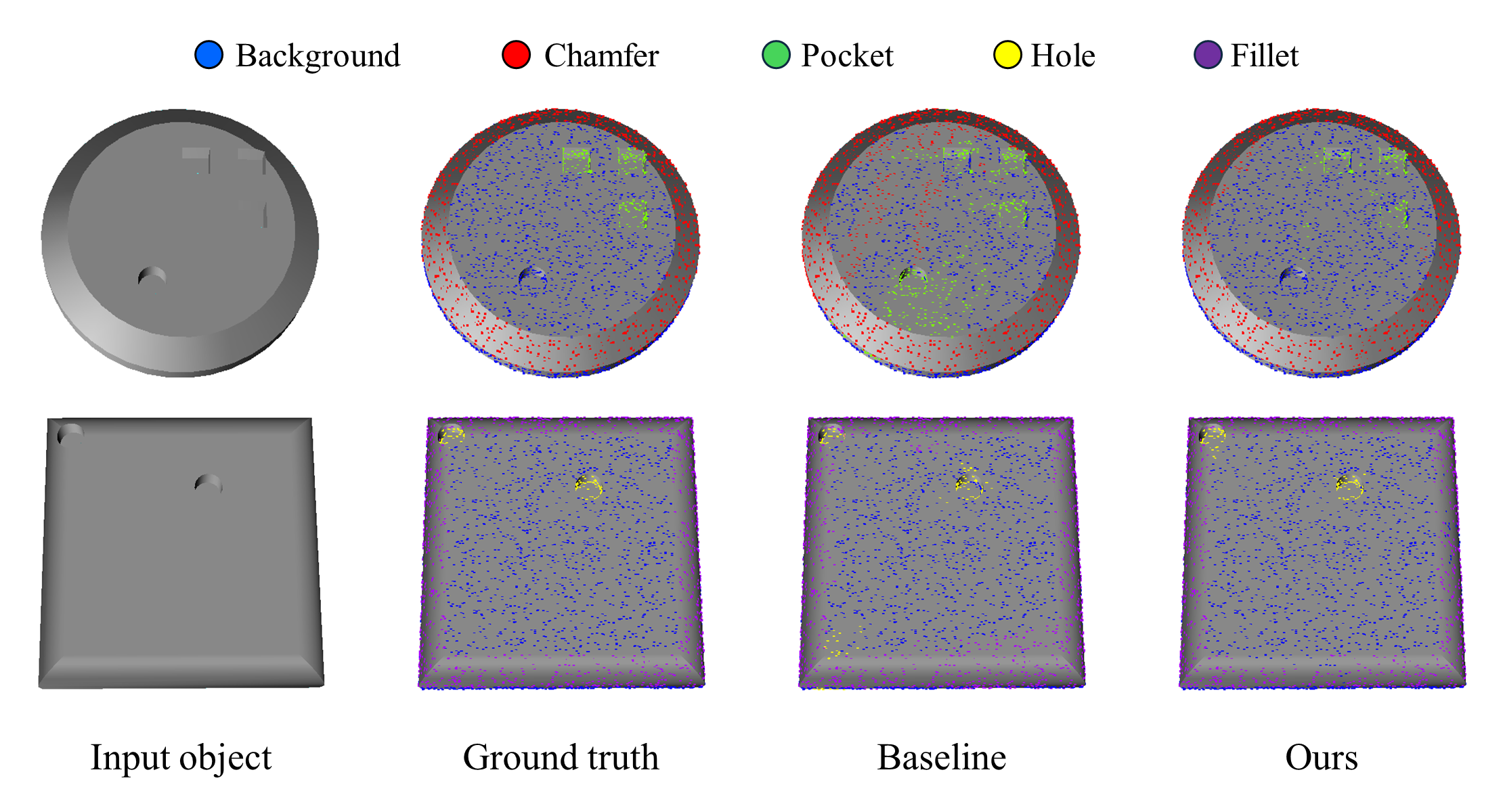}
  \caption{Visualized segmentation result and comparison on EIF 3D dataset.}
  \label{fig:pcr}
\end{figure}
\begin{figure}[t]
  \centering
  \includegraphics[width=1.0\linewidth]{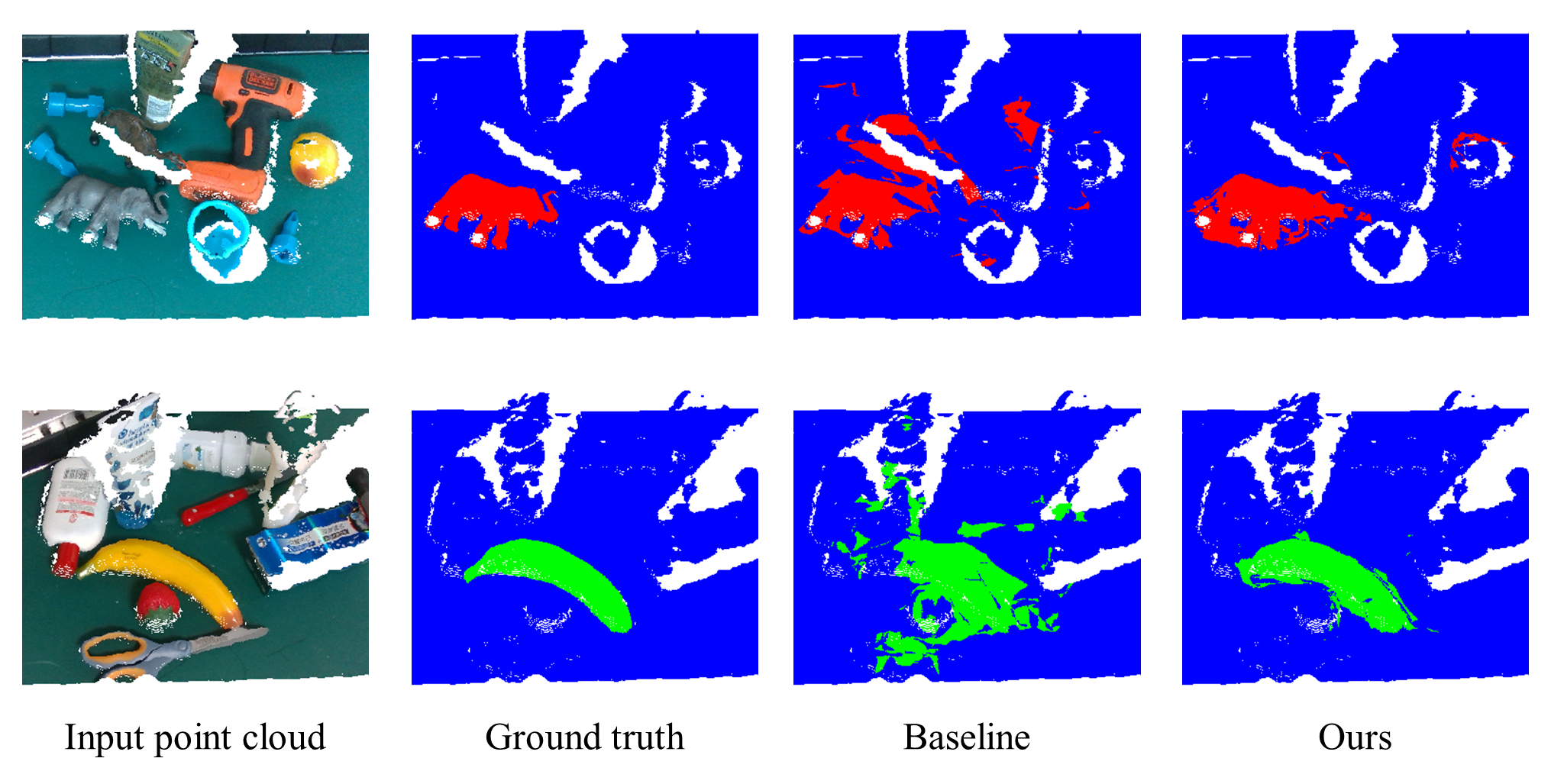}
  \caption{Visualized segmentation result and comparison on GraspNet dataset.}
  \label{fig:gcr}
\end{figure}
\begin{table*}[t]
    \tabcolsep=0.3cm
    \centering
    \begin{tabular}{l l l c}
    \toprule
    ID& Text Encoder & Prompt & Result\\
    \midrule
    \uppercase\expandafter{\romannumeral1}&CLIP~\cite{radford2021learning} & ``\textit{Chair}'' &73.08\\
    \uppercase\expandafter{\romannumeral2}&CLIP~\cite{radford2021learning} & ``\textit{Point clouds of a chair}'' &73.11\\
    \uppercase\expandafter{\romannumeral3}&CLIP~\cite{radford2021learning} & ``\textit{Point clouds of a chair in an office room}'' &73.08\\
    \uppercase\expandafter{\romannumeral4}&BGE-S~\cite{bge_m3} & ``\textit{Chair}'' &72.52\\
    \uppercase\expandafter{\romannumeral5}&BGE-B~\cite{bge_m3} & ``\textit{Chair}'' &72.90\\
    \uppercase\expandafter{\romannumeral5}&Qwen3-Embedding-0.6B~\cite{zhang2025qwen3} & ``\textit{Chair}'' &73.10\\
    \bottomrule
    \end{tabular}
    \caption{Ablation study of text encoders and prompts on S3DIS $S^0$ with 2-way 1-shot}
    \label{tb:appabl}
\end{table*}

\begin{table*}[t]
    \centering
    \begin{tabular}{l l c c c c c c}
    \toprule
 { Embed } & { Method } &2-way 1-shot&2-way 5-shot&$\Delta$ & 3-way 1-shot&3-way 5-shot&$\Delta$ \\
    \midrule 
word2vec & 3DGenZ & 29.07 & 31.65 & \textcolor{blue}{-30.84}
& 28.13 & 28.01 & \textcolor{blue}{-33.12}
\\
word2vec & PAPFZS3D & 54.72 & 58.94 & \textcolor{blue}{-4.37}
& 53.78 & 53.50 & \textcolor{blue}{-7.55}
\\
CLIP & PAPFZS3D & 56.12 & 60.65 & \textcolor{blue}{-2.81}
& 55.24 & 55.04 & \textcolor{blue}{-6.05}
\\
\rowcolor{lightblue!10}CLIP & \textbf{Ours} & \textbf{60.99} & \textbf{61.40} & - & \textbf{58.71} & \textbf{59.02} & - \\
    \bottomrule
    \end{tabular}
    \caption{Zero-Shot evaluation result on the ScanNet dataset using mean-IoU criteria (\%). The best result of each column is highlighted with \textbf{bold font}.}
    \label{tb:zeroshot_scannet}
\end{table*}

\subsection{Case Study on Practical Applications}
To showcase the tangible efficacy of our method in real-world scenarios, we construct few-shot datasets sourced from EIF 3D CAD~\cite{lee2022dataset,wang2022cam} models and GraspNet~\cite{fang2020graspnet} for evaluation. The EIF 3D dataset is an easily overfitted dataset that contains a few CAD models with manufacturing features such as pockets and chamfers. Semantic segmentation on the EIF 3D dataset can be treated as a simulation of Computer Aided Manufacturing (CAM) processes because the CAM usually requires a feature recognition process for the subsequent actions, such as tool path planning and tool controlling~\cite{wang2022cam}. Therefore, few-shot segmentation on the EIF 3D dataset demonstrates the potential of our method for deployment in the manufacturing industry. Specifically, Poisson sampling is adopted to convert the CAD model to point clouds, and 500 instances are manually labeled for testing. GraspNet~\cite{fang2020graspnet} is a public dataset for robotic grasping; we use the RGBD images and camera parameters to obtain the point cloud, thus constructing our few-shot inference episode.

Table~\ref{tb:phcfres} illustrates the quantitative result on the EIF 3D dataset. We can find that, although this dataset is small, our method can achieve satisfactory performance compared to other methods. Fig.~\ref{fig:pcr} is the visualization of our segmentation result. Fig.~\ref{fig:gcr} shows the visualized result and depicts that our method can provide better results than the baseline. Since the grasping task will benefit greatly from obtaining a prior mask of the target object, our method can help real-world robotic grasping.

\subsection{Prompts and Text Encoders}\label{sec:app_text}
Table~\ref{tb:appabl} presents the quantitative results of the corresponding experiments, using the ``chair'' class from the S3DIS dataset as an illustrative example to showcase the various prompt formulations. The table demonstrates that more specific prompts yield higher accuracy. However, if the prompt is too strict for a scene or a specific region, the performance will decrease. We also tested more specific prompts, such as ``A wooden, square-shaped chair with four metal legs placed in an office,” and found that the performance slightly decreased. This suggests that overly specific descriptions may hinder the generalization ability of text prototypes in the latent space. In other words, assigning unique text prompts to each instance can impair the model’s ability to learn category-level representations, leading to reduced generalization performance. Across different language models, the improvement remains marginal, indicating that as long as the model can produce distinguishable and robust representations for our prompts, it can effectively facilitate the entire pipeline. In practice, CLIP's multimodal training approach, which incorporates both text and visual data, gives it an advantage over encoders trained solely on textual information for our specific task; therefore, we select its text encoder as our language model. For fairness, all experiments in Section~\ref{sec:expresult} use the simplest prompt, e.g., ``chair''.

\begin{figure*}[!t]
    \centering
    \includegraphics[width=0.95\linewidth]{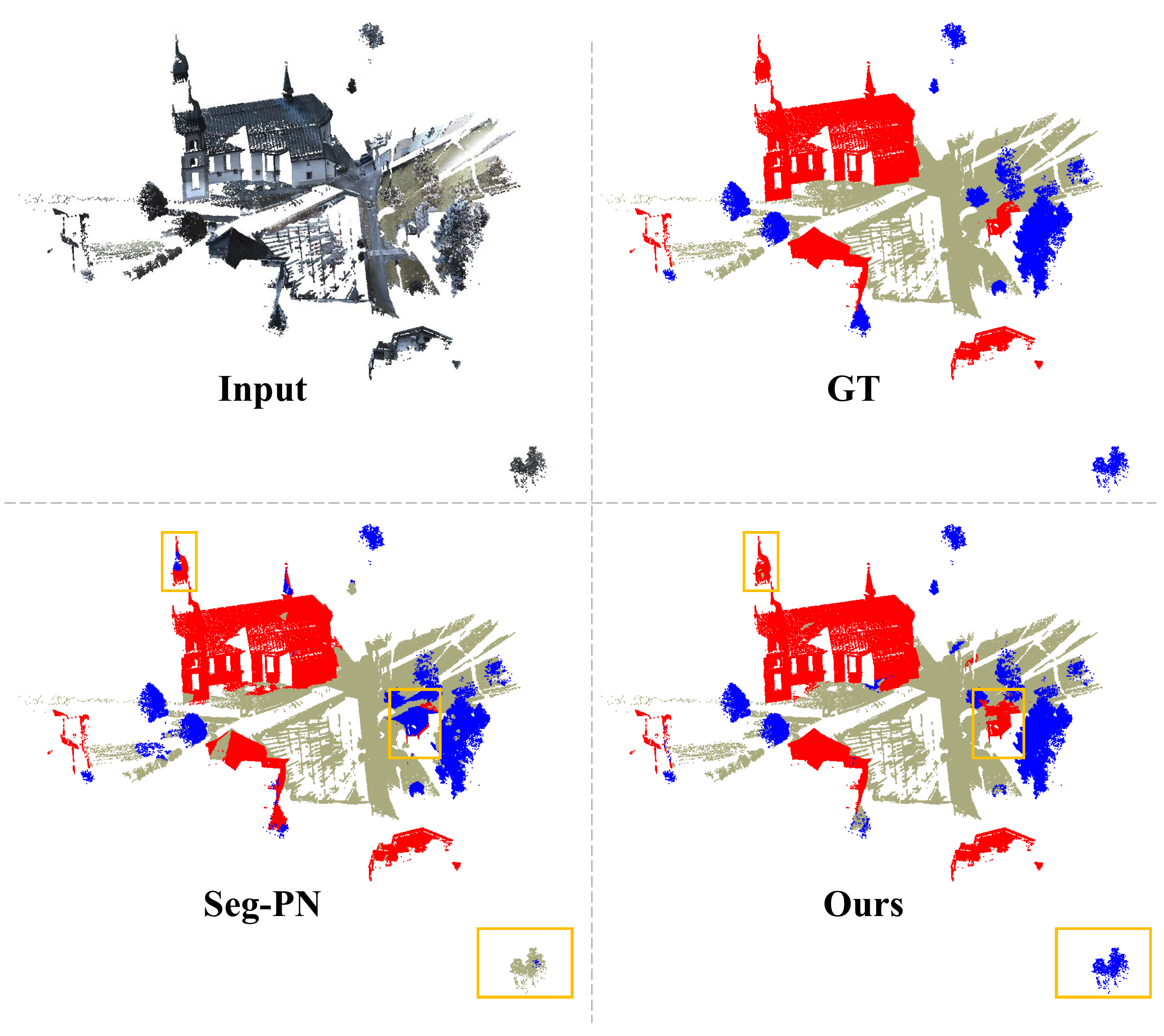}
    \caption{Visualization of segmentation result on Semantic3D dataset (2-way 1-shot). The red points represent buildings, the blue points denote high-vegetation areas, and the light grey points are labeled as background.}
    \label{fig:sem3d_res}
\end{figure*}
\subsection{Zero-Shot Experiment on ScanNet}
To complement the two mainstream benchmarks in 3D FS-SemSeg, we also evaluate the zero-shot performance of our model on the ScanNet~\cite{scannet} dataset. The experimental results are shown in Table~\ref{tb:zeroshot_scannet}. Compared to PAPFZS3D~\cite{PAPFZS3D} and 3DGenZ~\cite{3DGenZ}, our model achieves better performance, which shows the great potential of our proposed framework.

\begin{table}[t]
\centering
   \resizebox{\linewidth}{!}{ \begin{tabular}{c c c c c c c c}
    \toprule
     Model & \makecell{2-way\\ 1-shot} & \makecell{2-way\\ 5-shot} &$\Delta$  & \makecell{3-way\\ 1-shot} & \makecell{3-way\\ 5-shot}&$\Delta$\\
    \midrule
    AttMPTI & 42.27 &46.61&\textcolor{blue}{-12.98} & 31.75  & 33.46&\textcolor{blue}{-17.60}  \\
    PAPFZS3D & 48.42 & 49.91&\textcolor{blue}{-8.26}  & 43.07  & 45.18 &\textcolor{blue}{-6.08} \\
    Seg-PN &  51.58 & 54.71&\textcolor{blue}{-4.28}  &45.22  & 50.09 &\textcolor{blue}{-2.55}\\
    \rowcolor{lightblue!10}\textbf{Ours} & \textbf{54.49}	& \textbf{58.35} & - & \textbf{48.16} & \textbf{52.25}&- \\
    \bottomrule
    \end{tabular}}
    \caption{Quantitative result on the Semantic3D dataset.}
    \label{tb:extra}
\end{table}

\begin{table}[!ht]
\centering
   \resizebox{\linewidth}{!}{ \begin{tabular}{c c c c c c c c}
    \toprule
     Model & \makecell{2-way\\ 1-shot} & \makecell{2-way\\ 5-shot} &$\Delta$  & \makecell{3-way\\ 1-shot} & \makecell{3-way\\ 5-shot}&$\Delta$\\
    \midrule
    AttMPTI &47.42 & 51.56 &\textcolor{blue}{-8.87} & 38.95  & 41.13 &\textcolor{blue}{-12.60}  \\
    PAPFZS3D & 50.61 & 55.08 &\textcolor{blue}{-5.52}  &44.38  & 49.02 &\textcolor{blue}{-6.29} \\
    Seg-PN &  52.70 & 57.62 &\textcolor{blue}{-3.20}  & 46.65  & 50.93 &\textcolor{blue}{-4.21}\\
    \rowcolor{lightblue!10}\textbf{Ours} & \textbf{55.48}	& \textbf{61.24} & - & \textbf{50.80} & \textbf{55.19}&- \\
    \bottomrule
    \end{tabular}}
    \caption{Quantitative result on the SemanticKITTI dataset.}
    \label{tb:extra_kitti}
\end{table}
\subsection{Generalization to Outdoor Data}
For 3D FS-SemSeg, most of the research focuses on the indoor scanning environment. Such an environment is controllable (set up by humans) and can be less noisy. Noticed that our proposed method is not only appropriate for indoor scanning data but also able to deal with outdoor data. Semantic3D~\cite{hackel2017isprs} and SemanticKITTI~\cite{semantickitti} are well-known outdoor datasets for their large scene scale and precise annotation. For Semantic3D, we evaluate our model on its subset Semantic3D-8, which consists of 8 classes with over a billion points. We randomly select half of the classes (4 semantic categories as novel classes) and use the annotation of points belonging to the remaining classes for training. For SemanticKITTI~\cite{semantickitti}, we randomly select 200 scenes from the first 11 sequences, resulting in a total of 2200 scans. We then randomly choose half of the semantic categories for training, while the remaining categories are used for testing. Table~\ref{tb:extra} and Table~\ref{tb:extra_kitti} report the experimental results. Despite the lower numerical performance on these datasets compared to indoor data, our method consistently surpasses other baselines, highlighting its robustness. Figure~\ref{fig:sem3d_res} presents the visualized segmentation results on the Semantic3D-8~\cite{hackel2017isprs} dataset under the 2-way 1-shot setting. The input point cloud represents an outdoor scene characterized by complex geometry and significant noise. Compared to the baseline method (Seg-PN~\cite{segpn}), our approach demonstrates noticeably superior performance, particularly in high-frequency regions such as object boundaries, church steeples, and isolated structures. These results highlight the effectiveness of our method in preserving both low- and high-frequency information, as well as its robustness to noisy environments.

\end{document}